\documentclass[journal]{IEEEtran} 

\IEEEoverridecommandlockouts

\usepackage{amsmath}
\usepackage{amssymb}
\usepackage[geometry]{ifsym}
\usepackage{cite}
\usepackage{graphicx}
\newsavebox\mybox

\usepackage{ifsym}

\usepackage{caption}
\usepackage{subcaption}
\usepackage{array}
\usepackage[hidelinks]{hyperref}
\usepackage{enumitem}
\usepackage{dblfloatfix}

\usepackage{enumerate}
\usepackage{algorithmic}
\usepackage[ruled,linesnumbered,noend]{algorithm2e}

\SetArgSty{textnormal}
\SetKw{Continue}{continue}

\usepackage[normalem]{ulem}
\usepackage{color}
\usepackage[dvipsnames]{xcolor}

\newcommand{\APbinding}{AP_{\psi}}
\newcommand{\APltl}{AP_{\varphi}}
\newcommand{\ltlpsi}{LTL$^\psi$}

\newcommand{\buchi}{B{\"u}chi }

\newcommand{\PiTF}{\Pi_{F}}
\newcommand{\PiFT}{\Pi_{T}}
\newcommand{\PiexT}{\Pi_{exT}}
\newcommand{\PiexF}{\Pi_{exF}}

\newcommand{\PixT}{\Pi_{\emptyset T}}
\newcommand{\PixF}{\Pi_{\emptyset F}}

\newcommand{\PixexT}{\Pi_{\emptyset exT}}
\newcommand{\PixexF}{\Pi_{\emptyset exF}}

\newcommand{\sigIntOne}{\sigma_{11^{*}}}

\newcommand{\noninstFunc}{\mathcal{I}}
\newcommand{\buchiUpd}{\mathcal{B}_{upd}}
\newcommand{\prodUpd}{\mathcal{G}_{j,upd}}
\newcommand{\sigExT}{\sigma^{exT}}
\newcommand{\sigExF}{\sigma^{exF}}

\newcommand{\syntline}{\: | \:}
\newcommand{\X}{\bigcirc}
\newcommand{\G}{\Box}
\newcommand{\F}{\Diamond}
 
\usepackage{amsthm}
\theoremstyle{slanted}
\newtheorem{definition}{Definition}

\newtheorem{rem}{Remark}
\theoremstyle{remark}
\makeatletter
\let\NAT@parse\undefined
\makeatother
\usepackage{hyperref}  

\usepackage[normalem]{ulem}
\usepackage{color}
\usepackage{xcolor}

\newcommand{\new}[1]{{#1}}

\usepackage{hyperref}

\usepackage{xcolor} 
\usepackage{tikz}
\usetikzlibrary{fit,calc}

\colorlet{pink}{red!40}

\title{\bf
Continuous Execution of High-Level Collaborative Tasks for Heterogeneous Robot Teams}
\author{Amy Fang, Tenny Yin, Jiawei Lin and Hadas Kress-Gazit
\thanks{Sibley School of Mechanical and Aerospace Engineering, Cornell University, Ithaca, NY, 14853 USA. {\tt\small \{axf4, yy389, jnl77, hadaskg\}@cornell.edu}. This work is supported by the 
NDSEG Fellowship Program.
}}

\begin{document}
\maketitle
\thispagestyle{empty}
\pagestyle{empty}

\begin{abstract}
    We propose a control synthesis framework  for a heterogeneous multi-robot system to satisfy collaborative tasks, where actions may take varying duration of time to complete. We encode tasks using the discrete logic \ltlpsi, which uses the concept of bindings to interleave robot actions and express information about relationship between specific task requirements and robot assignments. We present a synthesis approach to automatically generate a teaming assignment and corresponding discrete behavior that is correct-by-construction for continuous execution, while also implementing synchronization policies to ensure collaborative portions of the task are satisfied. We demonstrate our approach on a physical multi-robot system.
\end{abstract}


\section{Introduction}

This paper presents a framework to generate decentralized controllers and synchronization signals, when needed, for a team of heterogeneous robots to satisfy a collaborative task. We consider tasks in which the user requires specific actions to be performed, but does not have requirements on the number or type of robots needed to satisfy a task. For example, in a warehouse environment, the user may want to move a pallet and then pick up a package from the pallet. Depending on the number of available robots and their capabilities, the robots may decide to collaborate to satisfy the task, or perform it by themselves. For instance, there may be a mobile manipulator that can perform both actions on its own, or, if it cannot handle a heavy pallet but there is a stationary manipulator that can, the robots will team up so that one moves the pallet and then the other picks up the package.  

In our previous work \cite{Fang2024}, we introduced the logic \ltlpsi \ that allows us to capture such tasks, and proposed a framework that transforms an \ltlpsi \ task to high-level controllers.
The logic uses \textit{bindings} to allow
users to specify the relationship between robots and parts of
the task without providing explicit assignments or constraints
on the number of robots required. In \cite{Fang2024}, the synthesized behavior assumes actions are discrete and treated as instantaneous. However, in physical systems, robots may execute actions with varying duration, such as picking up an object or navigating to an adjacent room. As a result, the synthesized behaviors may violate parts of the specification when executed by the robots. For example, consider a task that requires a surveillance robot to monitor a room only if other robots are present. In the symbolic controller, the two robots can both be outside the room, then immediately be in the room in the next step of their behavior. However, when the robots execute this desired behavior, it is extremely unlikely that they make the transition into the room at the exact same time - any mismatch of timing will cause the task to fail. 

In addition, under the instantaneous action assumption of~\cite{Fang2024}, tasks of the form ``all robots must enter the room at the same time" and ``at least one robot must enter the room" can be solved by assuming all robots with that binding move; however, under the varying action duration assumptions, tasks such as ``all robots must enter the room at the same time" are unrealistic and cannot be synthesized. We modify the synthesis algorithms in~\cite{Fang2024} to explicitly look for solutions of the form ``at least one robot" when such specifications are given, 

To address these issues, in this paper we define the semantics and synthesize symbolic correct-by-construction controllers for \ltlpsi\ tasks such that the continuous-time behavior of the robots is still guaranteed to satisfy the specification during execution.

\subsection{Related Work}
        

There is a wealth of literature in task assignment and
scheduling for multi-robot systems. 
Multi-robot task allocation problems are often treated as optimization problems \cite{Atay2006,Suslova2020,Tolmidis2013}, such as the multiple traveling salesman problem \cite{Cheikhrouhou2021} or the vehicle routing problem \cite{Braekers2016}. To mitigate the challenges of finding an exact solution, common approximate methods are used to solve the problem, such as learning algorithms, heuristics and meta-heuristics \cite{Poudel2021} \cite{Wu2018}, and contract net protocols \cite{Holvoet2007} \cite{Zhen2021}. Coalition formation algorithms allow robots to automatically form teams to execute a task. Approaches include swarm algorithms \cite{Xu2015} and multi-stage coalition formation \cite{Liu2016}, where the authors provide a two-stage approach to prune for a satisfying coalition.


The aforementioned approaches adopt a more generalized representation of tasks that are agnostic to the specific characteristics of each task, with the primary objective being to map agents to tasks. {Each task is abstracted with certain constraints associated with it, such as cost or number of agents required. While these methods  enable efficient task allocation algorithms, they do not account for the temporal dependencies and sequences of actions within the task itself.}
For describing temporally extended tasks, temporal logics can be useful;  they allow users to rigorously define temporally extended tasks that may require complex action sequences and constraints. These logics define tasks with symbolic abstractions of the given continuous system. For multi-robot systems, existing work has extended Linear Temporal Logic (LTL) \cite{Luo2022, Sahin2017, Chen2021} and Signal Temporal Logic (STL) \cite{Leahy2022} \cite{Gundana2021} to encode tasks that require multiple robots to execute.

To ensure the satisfaction of multi-robot temporal logic specifications in continuous time, a common approach is to encode the task in STL, which allows for discrete-time continuous signals. For example, in \cite{Buyukkocak2021}, the authors synthesize coordinated trajectories of a heterogeneous multi-robot team to satisfy a global task written in STL. They also introduce integral predicates to provide a quantitative metric for the cumulative progress of the tasks. The type of coordination that can be encoded only includes accumulation (e.g. data collection), as opposed to highly collaborative tasks that require strict synchronization. 
{To address strict synchronization constraints, the authors of \cite{Tumova2016} propose an event-based synchronization approach in which robots execute local LTL specifications and send synchronization requests when needed. This removes the necessity to synchronize at every discrete step, thus increasing efficiency. However, this work assumes that each robot is assigned a local specification \textit{a priori}, as opposed to collectively satisfying a global task.}

In \cite{Liu2023}, the authors address the problem of synthesizing controllers for a multi-agent system by extending Capability Temporal Logic (CaTL), which is a fragment of STL, to CaTL+. To ensure satisfiability, they encode two layers of logic, one for individual trajectories and one for the team trajectory. The controllers for each agent are then automatically synthesized by solving a centralized two-step optimization problem. For synchronization requirements, each task includes a counting proposition to indicate the timesteps in which synchronization must happen. The task explicitly encodes the number of capabilities required for each part of the task. In our work, we use \ltlpsi\ to define multi-agent tasks, which is an LTL-based grammar we first proposed in  \cite{Fang2024}. The grammar allows users to define tasks based on actions required instead of the number of agents/capabilities (e.g. “move the pallet and then pick up the package, irrespective of how many agents perform which actions"). 
In our work, we provide guarantees about the satisfaction of continuous controllers while using a discrete LTL-based logic. In this way, we can maintain the discrete abstraction of actions without any information about their specific timings.


To take into account the fact that actions may have varying durations, the authors in \cite{Raman2012} generate reactive hybrid controllers for LTL tasks such that the continuous behavior of the robots are safe. The authors abstract actions with timing constraints by using initiation and completion propositions, and add constraints in the specification for activating these propositions. The specification also includes constraints to ensure collision avoidance across robots. While this approach accounts for collision avoidance, it does not consider collaborative tasks and therefore also does not account for any necessary synchronization constraints.
\cite{Moarref2017} uses synchronization skeletons to coordinate a homogeneous swarm of robots to collectively satisfy a task. The task contains both a macro specification that describes the behavior of a group of robots, and a micro specification for individual robots to satisfy.
To ensure the behavior satisfies the specification in the continuous space (i.e. when robots are between two regions), the authors propose an iterative method of 
synthesizing labeled transition systems, then removing unsafe transitions and adding safety formulas until a valid labeled transition system is found. Rather than an iterative synthesize-then-check approach, our work aims to automatically abstract continuous actions and synthesize symbolic controllers that are already correct-by-construction for continuous execution.

\noindent \textbf{Contributions}: 
In this paper, we propose a framework for control synthesis for continuous execution of collaborative tasks, where actions may take varying duration of time to complete. Based on~\cite{Fang2024}, we use \ltlpsi to encode such tasks and present a synthesis approach to automatically generate a teaming assignment and corresponding symbolic behavior that is correct-by-construction during continuous execution, while also ensuring synchronization requirements are satisfied for collaborative portions of the task. We demonstrate our approach on a physical multi-robot system.



        



\section{Preliminaries}

\subsection{Linear Temporal Logic}\label{sec:ltl}

LTL formulas are constructed from atomic propositions $AP$, where $\pi \in AP$ are Boolean variables \cite{EMERSON1990}. The atomic propositions represent an abstraction of robot actions. For example, $camera$ captures a robot taking a picture. 

\textbf{Syntax: } We use the following syntax to define an LTL formula:
\[
    \varphi ::=  \pi  \ |  \ \neg\varphi  \ | \ \varphi \vee \varphi \ | \ \bigcirc \varphi \ | \ \varphi \ \mathcal{U} \  \varphi
\]

\noindent where $\neg$ and $\vee$ are Boolean operators (``not" and ``or", respectively), and $\bigcirc$ and  $\mathcal{U}$ (``next" and ``until", respectively) are temporal operators.
Using these, we can also define conjunction $\varphi \wedge \varphi$, implication $\varphi \Rightarrow \varphi$, eventually $\Diamond \varphi = \text{True} \ \mathcal{U} \ \varphi $, and always $\Box \varphi=\neg\Diamond \neg\varphi$.

\textbf{Semantics:} The semantics of an LTL formula $\varphi$ are defined over an infinite trace $\sigma = \sigma(0)\sigma(1)\sigma(2)...$, where $\sigma(i)$ is the set of $AP$ that are true at step $i$. We use $\sigma \models \varphi$ to denote that the trace $\sigma$ satisfies LTL formula $\varphi$.

Intuitively, $\bigcirc \varphi$ is satisfied if $\varphi$ is satisfied in the next step of the sequence;  $\Diamond \varphi$ is satisfied if $\varphi$ is true at some step in the sequence; $\Box \varphi$ is satisfied if $\varphi$ is true at every step in $\sigma$; $\varphi_1 \ \mathcal{U} \ \varphi_2$ is satisfied if $\varphi_1$ remains true until $\varphi_2$ becomes true. See \cite{EMERSON1990} for the full semantics. 

\subsection{\buchi Automata} \label{sec:buchi_standard}

An LTL formula $\varphi$ can be translated into a Nondeterministic \buchi Automaton that accepts infinite traces if and only if they satisfy $\varphi$. A \buchi automaton is a tuple $\mathcal{B}= (Z, z_0, \Sigma_{\mathcal{B}}, \delta_{\mathcal{B}}, F)$, where $Z$ is the set of states, $z_0 \in Z$ is the initial state, $\Sigma_{\mathcal{B}}$ is the input alphabet, $\delta_{\mathcal{B}}: Z \times \Sigma_{\mathcal{B}} \times Z$ is the transition relation, and $F \subseteq Z$ is a set of accepting states.
An infinite run of $\mathcal{B}$ over a word $\sigma = \sigma_1 \sigma_2 \sigma_3 ... \in \Sigma_{\mathcal{B}}$ is an infinite sequence of states $z = z_0 z_1 z_2...$ such that $(z_{i-1}, \sigma_i, z_{i}) \in \delta_{\mathcal{B}}$. A run is accepting if and only if Inf($z$) $\cap \ F \neq \emptyset$, where Inf($z$) is the set of states that appear in $z$ infinitely often \cite{Baier2008}.

\section{Definitions}

\subsection{Actions}

We categorize the actions the robots can execute into two types, \textit{instantaneous} and \textit{non-instantaneous}. For example, the action of taking a picture is considered instantaneous; in contrast, the action of moving between rooms is non-instantaneous.

Each action is abstracted as an atomic proposition (AP) $\pi$, and its corresponding completion proposition, $\pi_c$. $\pi$ is true at $\sigma(i)$ iff the action associated with $\pi$ is being executed at position $i$ in the trace; $\pi_c \in AP$ is true at $\sigma(i)$ iff the action associated with $\pi$ has been completed at position $i$ in the trace. For example, $roomA$ indicates that the robot is currently moving towards room A; $roomA_c$ indicates the robot has finished executing the action $roomA$ and is currently in room A.

The set of all $AP$ is composed of the subsets $AP_{inst}$ and $AP_{non-inst}$, which are the sets of propositions representing instantaneous actions and non-instantaneous actions, respectively. Formally, an action is instantaneous iff $T(\pi, \pi_c) = 0$, where $T$ is the timing function that outputs the amount of time it takes for the action to be executed; thus, $\pi_c \in AP_{inst}$. Similarly, an action is considered non-instantaneous iff $T(\pi, \pi_c) \neq 0$. For these types of actions, $\pi_c \in AP_{non-inst}$.  For example, for the action of beeping, $T(beep, beep_c) = 0$ and therefore $beep_c \in AP_{inst}$. In contrast,  $T(roomA, roomA_c) \neq 0$ (i.e. moving into room A takes time), so $roomA_c \in AP_{non-inst}$.

Note that, for synchronization purposes, we assume that robots can coordinate such that they can execute an action simultaneously, so all $\pi$ are in the set of $AP_{inst}$. For example, $roomA_c \in AP_{non-inst}$ but $roomA \in AP_{inst}$; $beep, beep_c \in AP_{inst}$ (note that, for instantaneous actions, $\pi$ and $\pi_c$ have equivalent meanings).

\subsection{Robot Model} \label{sec:robot_model}

Based on our prior work~\cite{Fang2022}, each robot $j$ is modeled based on its set of capabilities, $\Lambda_j =  \{\lambda_1, ..., \lambda_k\}$. Each capability is a weighted transition system $\mathcal{\lambda} = (X, x_0, AP, \Delta, \mathcal{L}, \mathcal{W})$, where $X$ is a finite set of states, $x_0 \in X$ is the initial state, $AP$ is the set of atomic propositions representing the actions the capability can execute, $\Delta \subseteq X \times X$ is a transition relation,  $\mathcal{L}: X \rightarrow 2^{AP}$ is the labeling function, and $\mathcal{W}: \Delta \rightarrow \mathbb{R}_{\ge 0}$ is the cost function assigning a weight to each transition. The cost function is predefined by the user and can represent execution time, power usage, etc.

\begin{figure}
    \centering            \includegraphics[width=0.95\columnwidth]{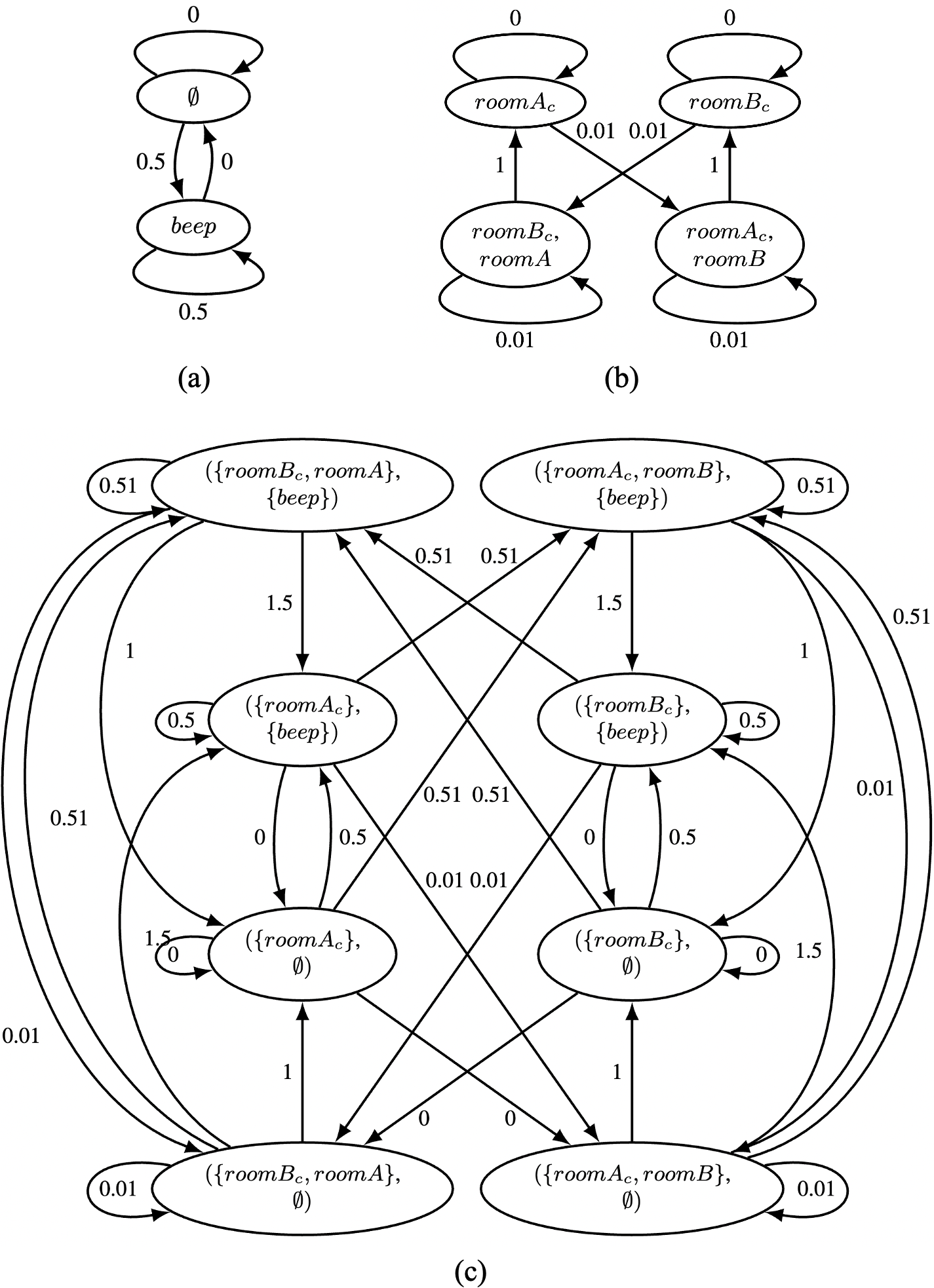}
\caption{Partial model of robot $A_{blue}$ (see Sec. \ref{sec:setup}): (a) $\lambda_{\mathit{beep}}$ (b) $\lambda_{\mathit{motion}}$ (c) $A_{blue}$
}
\label{fig:robot}
\end{figure}

A robot model $A_j$ is the cross product of its capabilities: $A_j = \lambda_1 \times... \times \lambda_k$ such that $A_j=(S, s_0, AP_j, \gamma, L, W)$. $S = X_1 \times ... \times X_k$ is the set of states, $s_0 \in S$ is the initial state, $AP_j = \bigcup_{i=1}^k AP_i$
is the set of propositions, $\gamma \subseteq S \times S$ is the transition relation, $L:S \rightarrow 2^{AP_j}$ is the labeling function, and $W: \gamma \rightarrow \mathbb{R}_{\ge 0}$ is the cost function (a function of $\mathcal{W}_i$). More details on the full definition can be found in \cite{Fang2022}. See Fig. \ref{fig:robot} for an example of a robot model. 

\subsection{Task Grammar - \ltlpsi} 

We first introduced \ltlpsi in \cite{Fang2024}. In this work, because we are removing the assumption of instantaneous behavior, the ``next" operator is no longer meaningful therfore we omit it. 
The task grammar for \ltlpsi includes atomic propositions that abstract robot action, logical and temporal operators, as in LTL, and bindings that relate actions to specific robots; any action labeled with a certain binding must be satisfied by all the robot(s) assigned that binding. 

We define a task $\varphi^{\psi}$ recursively over LTL and binding formulas.
\begin{flalign}
    \psi &:= \rho \syntline \psi_1 \vee \psi_2 \syntline \psi_1 \wedge \psi_2 \\
     \varphi &:=  \pi  \ |  \ \neg\varphi  \ | \ \varphi \vee \varphi \ | \ \varphi \ \mathcal{U} \  \varphi \\
     \varphi^{\psi} \!&:= \varphi^{\psi} \! \syntline
     \neg(\varphi^{\psi}) \syntline
    \! \varphi_1^{\psi_1} \!\! \wedge \! \varphi_2^{\psi_2} | \
    \! \varphi_1^{\psi_1} \!\! \vee \! \varphi_2^{\psi_2} \! \syntline 
     \varphi_1^{\psi_1} \mathcal{U} \varphi_2^{\psi_2} \! \syntline 
     \G \varphi^{\psi}\hspace{-3mm}
\end{flalign} 


\noindent where $\psi$, the \textit{binding formula}, is a Boolean formula excluding negation over the binding propositions $\rho\in\APbinding$, and $\varphi$ is an LTL formula consisting of the set of propositions $AP_{\varphi}$.

\textbf{Semantics: }   
The semantics of an \ltlpsi \ formula $\varphi^{\psi}$ are defined over the team trace $\sigma = \sigma_1\sigma_2...\sigma_n$, where each $\sigma_j$ is the trace for robot $j$; and the set of binding assignments $R = \{r_{1}, r_{2},..., r_{n}\}$, where $r_j \in R $ is the set of bindings in $\APbinding$ that are assigned to robot $j$. A robot's binding assignment does not change throughout the task execution (i.e. $R$ remains the same). 
For example, $r_{1} = \{2\}, r_{2} = \{1,3\}$ indicates that robot 1 is assigned binding 2, and robot 2 is assigned bindings 1 and 3. 

Given a set of binding propositions $\APbinding$, $\zeta: \psi \rightarrow 2^{2^{\APbinding}}$ outputs all possible combinations of $\rho \in \APbinding$ that satisfy $\psi$. For example, $\zeta \bigl( (1 \wedge 2) \vee 3 \bigr) = \{ \{1,2\},\{3\}, \{1,2,3\}\}$. We say that a team of robots satisfies the formula $\varphi^{\psi}$ if and only if a binding assignment exists in $\zeta(\psi)$ such that all the bindings are assigned to (at least one) robot, and the behavior of all robots with these binding assignments satisfy $\varphi$. A robot may be assigned to multiple bindings, and a binding may be assigned to multiple robots. 

Formally, the semantics are defined recursively as follows:

\begin{itemize}
    \item $(\sigma(i), R)\! \models \varphi^{\psi}$ iff $\exists K \in \zeta(\psi)$ s.t. ($ K \subseteq \bigcup\limits_{p=1}^{n} r_{p}$) and ($\forall j$ s.t. $K \cap r_j \neq \emptyset$, $\sigma_j(i) \models \varphi$)
    
    \item $(\sigma(i), R)\! \models (\neg \varphi)^{\psi}$ iff $\exists K \in \zeta(\psi)$ s.t. ($ K \subseteq \bigcup\limits_{p=1}^{n} r_{p}$) and ($\forall j$ s.t. $K \cap r_j \neq \emptyset$, $\sigma_j(i) \not  \models \varphi$)

    \item $(\sigma(i), R)\! \models \neg(\varphi^{\psi})$ iff $\exists K \in \zeta(\psi)$ s.t. ($ K \subseteq \bigcup\limits_{p=1}^{n} r_{p}$) and ($\exists j$ s.t. $K \cap r_j \neq \emptyset$, $\sigma_j(i) \not\models \varphi$)
    
    \item $(\sigma(i),\!R)\!\!\models\!\! \varphi_1^{\psi_1} \!\wedge\varphi_2^{\psi_2}$ iff $(\sigma(i),\!R)\!\!\models \! \varphi_1^{\psi_1}$and $(\sigma(i),\!R)\! \models\!\varphi_2^{\psi_2}$
    
    \item $(\sigma(i),\!R)\! \models\! \varphi_1^{\psi_1}\!\vee\! \varphi_2^{\psi_2}$ iff $(\sigma(i),\!R)\! \models\!\varphi_1^{\psi_1}$or $(\sigma(i), R)\! \models \varphi_2^{\psi_2}$
    

    
    \item $(\sigma(i), R) \models \X \varphi^{\psi}$ iff $\sigma(i+1), R \models \varphi^{\psi}$
    

    \item $(\sigma(i), R)\! \models \varphi_1^{\psi_1} \mathcal{U} \varphi_2^{\psi_2}$ iff $\exists \ell \geq i$ s.t. $(\sigma(\ell), R) \models \varphi_2^{\psi_2}$ and $\forall i \leq k < \ell, (\sigma(k), R) \models \varphi_1^{\psi_1}$
    
      \item $(\sigma(i), R) \models \G \varphi^{\psi}$ iff $\forall \ell>i, (\sigma(\ell), R) \models \varphi^{\psi}$


\end{itemize}

\begin{rem}
    The notation $\neg \varphi^\psi$ and $(\neg \varphi)^\psi$ are equivalent. For example, $\neg roomB^1 \triangleq (\neg roomB)^1$.
\end{rem}

\begin{rem}
    There is an important distinction between $(\neg \varphi)^{\psi}$ and $\neg (\varphi^{\psi})$. $(\neg \varphi)^{\psi}$ requires that the traces of all robots with binding assignments that satisfy $\psi$,  satisfy $\neg \varphi$. $\neg (\varphi^{\psi})$ requires the formula $\varphi^{\psi}$ to be violated; at least one robot's trace with a binding assignment that satisfies $\psi$ does not satisfy $\varphi$ (equivalently, satisfies $\neg\varphi$).
\end{rem}

\begin{rem}
    Unique to \ltlpsi\ is the ability to encode constraints either on all robots or on at least one robot.
    $\varphi^\psi$ captures ``all robots assigned a binding in $K \in \zeta(\psi)$ must satisfy $\varphi$''; 
    $\neg(\neg \varphi^{\psi})$ captures ``at least one robot assigned a binding in $K \in \zeta(\psi)$ satisfies $\varphi$''. This allows the same binding to be assigned to multiple robots, but only one of those robots needs to satisfy $\varphi$. This can be especially valuable when encoding safety requirements; for example, the formula $\neg (\neg roomB^1) \Rightarrow (roomB \wedge camera)^2$ captures ``if any robot assigned binding 1 is in room B, all robots assigned binding 2 must take a picture of the room.'' 

\end{rem}

\noindent \textit{Example: Task 1} 

\begin{figure*}[h!]
    \centering 
    \includegraphics[width=0.95\textwidth]{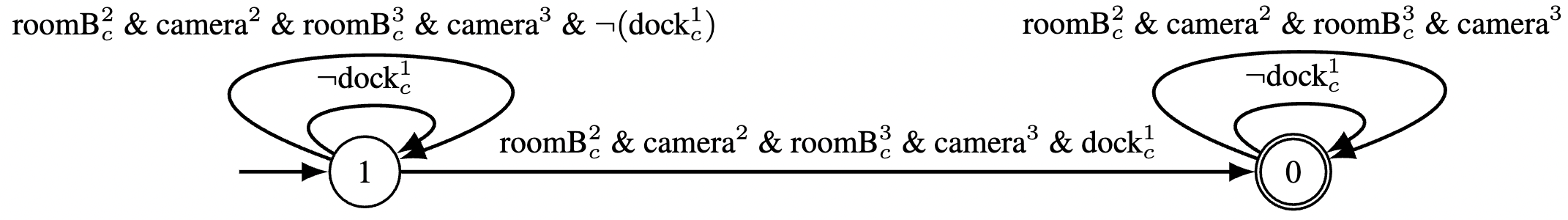}
    \caption{\buchi automaton for Task 1}
    \label{fig:buchi}
\end{figure*}


\begin{align} \label{ex:ex1}
\varphi^{\psi} = \F(dock_c^{1}) \wedge \G(\neg(\neg dock_c^{1}) \rightarrow (roomB_c \wedge camera)^{2 \wedge 3})
\end{align} 

In English, the specification captures ``All robot(s) assigned binding 1 must eventually be at the dock. Anytime one of these robots is at the dock, all robot(s) assigned bindings 2 and 3 must be in room B taking pictures." The second part of the task is a safety constraint that ensures that the dock is monitored anytime a robot assigned binding 1 is in that room.

\section{Problem Statement} \label{sec:setup}

Given $n$ heterogeneous robots $A = \{A_1,..., A_n\}$ and a task $\varphi^\psi$ in \ltlpsi, find a team of robots $\hat{A} \subseteq A$, their binding assignments $R_{\hat{A}}$, and synthesize behavior $\sigma_j$ for each robot such that $(\sigma(0), R_{\hat{A}}) \models \varphi^\psi$ and can by design be implemented in continuous execution (i.e., the physical execution of $\sigma_j$ does not violate any part of $\varphi$). Similar to our prior work, this behavior includes synchronization constraints implicitly encoded in the task for robots to satisfy the necessary collaborative actions. 

We assume that the system is nonreactive; robots cannot change their behavior at runtime. We also assume that every state in both the \buchi automaton of the task $\varphi^\psi$ and the robot model has a self-transition (i.e. the robot can wait in any state).

\begin{figure}[h!]
    \centering 
    \includegraphics[width=0.5\columnwidth]{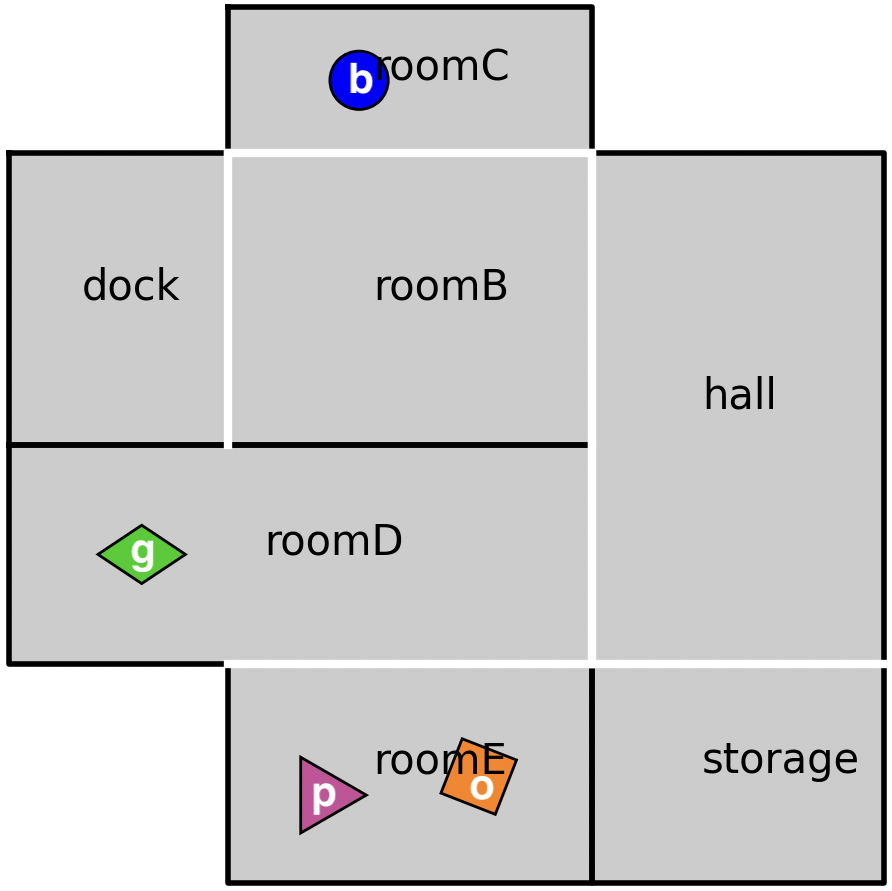}
    \caption{Environment and robot setup}
    \label{fig:setup}
\end{figure}

\noindent \textit{Example:}  
Let there be four robots $A = \{A_{green}$, $A_{blue}$, $A_{orange}$, $A_{pink}\}$ in a warehouse environment (Fig. \ref{fig:setup}).
The set of all capabilities is $\Lambda = \{\lambda_{mot}, \lambda_{arm}, \lambda_{beep}, \lambda_{cam}, \lambda_{scan}\}$, where $\lambda_{mot}$ is the motion model representing how the robots can move through the environment. $\lambda_{arm}$ represents a robotic manipulator; the arm can pick up and drop off packages, as well as push boxes.
$\lambda_{beep}$, $\lambda_{cam}$, and $\lambda_{scan}$ are a robot's ability to beep, take a picture, and scan, respectively. $\lambda_{mot}$ and $\lambda_{arm}$ execute non-instantaneous actions; the remaining capabilities contain only instantaneous actions.

The robots' capabilities and labels on their initial states are:

\noindent
$\Lambda_{green} = \{\lambda_{\mathit{mot}}, \lambda_{arm}, \lambda_{camera}\}, L(s_0) = \{roomD_c\}$ \\
$\Lambda_{blue} = \{\lambda_{\mathit{mot}}, \lambda_{beep}\}, L(s_0) = \{roomC_c\}$ \\
$\Lambda_{orange} = \{\lambda_{\mathit{mot}}, \lambda_{arm}\},L(s_0) = \{roomE_c\}$ \\
$\Lambda_{pink} = \{\lambda_{\mathit{mot}},  \lambda_{beep},\lambda_{cam}, \lambda_{scan}\}$, $L(s_0) = \{roomE_c\}$

The goal is to construct a team, i.e. a subset of these robots, and corresponding binding assignments for the robots to satisfy the task captured as the \ltlpsi specification provided in Eq. \ref{ex:ex1}. 
{We assume that the robots have low-level controllers that ensure collision avoidance and coordination when multiple robots execute a capability together (e.g. pushing a cart).}

\section{Approach}

To find a teaming assignment and synthesize the corresponding synchronization and control that is guaranteed to satisfy the task even when actions have varying execution times, we start with an approach similar to our prior work~\cite{Fang2024}: first, we automatically generate a \buchi automaton $\mathcal{B}$ for the task $\varphi^\psi$ (Sec. \ref{sec:buchi_ltlpsi}). Each robot $A_j$ then constructs a product automaton $\mathcal{G}_j=A_j \times \mathcal{B}$  (Sec. \ref{sec:prod_aut}). The novelty in our current work is our approach to finding a team of robots to satisfy the task. In prior work, we used a depth first search (DFS) algorithm to find both a team of robots and a trace through $\mathcal{B}$ to an accepting cycle. For every transition in the trace, every proposition in $\APbinding$ is assigned to at least one robot in the team. In this paper, we further ensure that the varying time durations of the actions do not result in violations of any part of the task. To do so, we update the \buchi automaton as the DFS progresses to include intermediate states that enforce the necessary ordering of actions without violating the original specification (Sec. \ref{sec:buchi_int}). As a result, we also update a robot's product automaton to reflect the changes in the \buchi automaton (Sec. \ref{sec:update_prod_aut}). The entire DFS framework is outlined in Sec. \ref{sec:DFS}.

Once an accepting trace and corresponding robot team is found, each robot synthesizes its own behavior that allows for parallel execution. However, some parts of the task may require collaboration, in which case the robots must synchronize. We create a synchronization policy where each robot communicates to others when they need to synchronize and waits for other robots before executing the collaborative portions of their behavior (Sec. \ref{sec:sync}). 

\subsection{\buchi Automaton for an \ltlpsi\ Formula}\label{sec:buchi_ltlpsi}
To create a \buchi automaton from an \ltlpsi\ specification, we first modify the specification to constrain binding propositions solely to individual atomic propositions $\pi \in AP_\varphi$ (i.e. the formula contains only $\pi^\rho$). For example, the formula $(\F (pickup \wedge roomA_c))^{1 \vee 2}$ is rewritten as $(\F (pickup^1 \wedge roomA_c^1) \vee (\F(pickup^2 \wedge roomA_c^2))$. 

Using $\APltl$ and $\APbinding$, we define $AP_{\varphi}^{\psi}$, where $\forall \pi \in \APltl$ and $\forall \rho \in \APbinding$, $\pi^\rho \in AP_{\varphi}^{\psi}$. We automatically create the \buchi automaton from these propositions using Spot \cite{spot}. 

Prior to control synthesis, we first convert the transitions in the \buchi automaton, labeled with Boolean formulas, into disjunctive normal form (DNF). Subsequently, we substitute the transition labeled with a DNF formula containing $\ell$ conjunctive clauses with $\ell$ transitions between the same states. The label for each of these transitions is a distinct conjunction of the original label. 

When constructing a \buchi automaton from an LTL formula $\varphi$, $\sigma \in\Sigma_{\mathcal{B}}$ are Boolean formulas over $AP_\varphi$, the atomic propositions that appear in $\varphi$. {In this work, for a \buchi automaton of an \ltlpsi\ formula, $\Sigma_{\mathcal{B}} = 2^{AP_{\varphi}^{\psi}}\times 2^{AP_{\varphi}^{\psi}}\times 2^{AP_{\varphi}^{\psi}}\times 2^{AP_{\varphi}^{\psi}}$, where $\sigma=(\sigma^T,\sigExT,\sigma^F, \sigExF)\in\Sigma_{\mathcal{B}}$ contains the set of propositions $\pi^\rho$ that must be true/false for all robots ($\sigma^T$ and $\sigma^F$), called \textit{for all} propositions, and the set of propositions $\pi^\rho$ that must be true/false for at least one robot ($\sigExT$ and $\sigExF$), called \textit{there exists} propositions. $\sigExT$ are the set of propositions in which $\neg(\neg \pi^\rho)$ is true; $\sigExF$ are the set of propositions in which $\neg(\pi^\rho)$ is true. Any proposition that does not appear in $\sigma$ can maintain any truth value.}

{We modify the definition of $\sigma$ compared to our prior work~\cite{Fang2024}, where we assumed instantaneous robot actions; there, $\sigma = (\sigma_T, \sigma_F)$. While~\cite{Fang2024} also had \textit{there exists} propositions, in practice the resulting behavior of \textit{there exists} and \textit{for all} propositions was the same. This is because, for a \textit{there exists} proposition $\pi^\rho$, only one robot assigned binding $\rho$ needs to execute $\pi^\rho$. However, since the robots do not coordinate, if there are multiple robots assigned $\rho$, they all satisfy $\pi^\rho$ (satisfying \textit{for all} propositions also satisfies \textit{there exists} propositions). Thus, it was unnecessary in~\cite{Fang2024} to differentiate between these two types of propositions. This is not true for time varying actions; there are scenarios in which a \textit{there exists} proposition is satisfied but a \textit{for all} proposition is not.} 

\textit{Example.} Consider the task defined in Eq. \ref{ex:ex1}. Supposed we have two robots with 
$r_{green} = \{1\}, r_{blue} = \{2,3\}$. Our prior approach may output a discrete teaming plan in which the blue robot enters the dock at exactly the same time the green robot enters room B. However, the physical execution of this plan may violate the specification; although the robots can begin moving towards their respective areas at the same time, the green robot may enter the dock before the blue robot enters room B, thus violating the safety constraint. 


\subsection{Constructing the Product Automaton} \label{sec:prod_aut}

To formally define the product automaton, we first modify the functions introduced in \cite{Fang2024}. 
For illustration, we will use the \buchi automaton for task 1 as our example (Fig. \ref{fig:buchi}). Let the outer self-transition on state 1 be $\sigma_{11, o}$. $\sigma_{11, o} = (\sigma_{11, o}^T,\sigExT_{11, o},\sigma_{11, o}^F, \sigExF_{11, o}) = (\{roomB_c^2, camera^2, roomB_c^3, camera^3\}, \emptyset, \emptyset, \{dock_c^1\})$.


\begin{definition}[Binding Transition Function]

$\mathfrak{B}_\sigma: \Sigma_{\mathcal{B}} \rightarrow 2^{ AP_\psi}$ such that for $\sigma{=(\sigma^T,\sigExT,\sigma^F, \sigExF)} \in \Sigma_{\mathcal{B}},  \mathfrak{B}_\sigma(\sigma) \subseteq \APbinding$ is the set $\{\rho \in\APbinding \ | \ \exists \pi^\rho \in {\sigma^T \cup \sigExT \cup \sigma^F \cup \sigExF\}}$.
\end{definition}

$\mathfrak{B}_\sigma(\sigma)$ outputs the set of bindings that appear in a \buchi transition  label $\sigma$. For example, 
$\mathfrak{B}_\sigma(\sigma_{11, o}) = \{1,2,3\}$.

\begin{definition}[Binding Set Function]
$\mathfrak{B}_\Pi: 2^{AP_{\varphi}^\psi} \rightarrow 2^{ AP_\psi}$ such that for $\Pi \subseteq AP_{\varphi}^\psi, \mathfrak{B}_\Pi(\Pi) \subseteq \APbinding$ is the set $\{\rho \in\APbinding \ | \ \exists \pi^\rho \in {\Pi\}}$.
\end{definition}

$\mathfrak{B}_\Pi(\Pi)$ outputs the set of bindings in a given set of \ltlpsi\ propositions $\Pi$. For example, for $\Pi = \{roomB_c^2, camera^2, roomB_c^3, camera^3, dock_c^1\}$, $\mathfrak{B}_\Pi(\Pi) = \{1,2,3\}$; for $\Pi =  \{dock_c^1\}$, $\mathfrak{B}_\Pi(\Pi) = \{1\}$.

\begin{definition}[Capability  Function]
$\mathfrak{C}: \Sigma_{\mathcal{B}} \times \APbinding \rightarrow 2^{\APltl} \times 2^{\APltl} \times 2^{\APltl} \times 2^{\APltl}$ such that for $(\sigma^T,\sigExT,\sigma^F, \sigExF) \in \Sigma_{\mathcal{B}}, \rho \in \APbinding$, $  \mathfrak{C}(\sigma, \rho) = (C_T, C_{exT}, C_F, C_{exF})$, where for $k \in \{T, exT, F, exF\}$, $C_k = \{\pi \in\APltl \ | \ \exists \pi^\rho \in \sigma^k\}$.
\end{definition}

$C_T$ and $C_F$ are the sets of \textit{for all} propositions that are True/False and appear with binding $\rho$ in label $\sigma$ of a \buchi transition and must be True/False for all robots assigned binding $\rho$; $C_{exT}$ and $C_{exF}$ are defined similarly but for \textit{there exists} propositions with binding $\rho$. For example, for $\sigma_{11, o} = (\{roomB_c^2, camera^2, roomB_c^3, camera^3\}, \emptyset, \emptyset, \{dock_c^1\})$, $\mathfrak{C}(\sigma_{11, o}, 2) = (\{roomB_c, camera\}, \emptyset, \emptyset, \emptyset)$ (for binding 2, the relevant propositions are $roomB_c$ and $camera$, both of which must be true and are \textit{for all} propositions, i.e. the set $\sigma^T$), and $\mathfrak{C}(\sigma_{11, o}, 1) = (\emptyset, \emptyset, \emptyset, \{dock_c\})$ ($dock_c$ is the only proposition with the binding 1 and is  in $\sigExF$).

\begin{definition}[Binding Assignment Function]
Given a robot model $A_j$ and a label 
$\sigma = (\sigma^T,\sigExT,\sigma^F, \sigExF)$, 
$\mathfrak{R}(s, \sigma, s') = \{r \in 2^{\APbinding} \setminus \emptyset \ | \ \forall \rho \in r, (C_T, C_{exT}, C_F, C_{exF})= \mathfrak{C}(\sigma, \rho),  \bigcup_{\rho \in r} (C_T \cup C_{exT}) \subseteq L(s')$ and $\bigcup_{\rho \in r} (C_F \cup C_{exF}) \cap L(s') = \emptyset \}$.
\end{definition}

To synthesize behavior for a robot, we find the minimum cost accepting trace in its product automaton $\mathcal{G}_j = A_j \times \mathcal{B}$, where $A_j = (S,s_0, AP_j, \gamma,L,W)$ is the robot model, and $\mathcal{B}= (Z,  z_0, \Sigma_{\mathcal{B}}, \delta_{\mathcal{B}}, F)$ is the \buchi automaton.

Given a transition $(s,\sigma, s')$, $\mathfrak{R}$ generates all possible combinations of binding propositions that can be assigned to a robot to satisfy $\sigma$. A robot can be assigned binding $\rho$ if and only if two conditions hold on the robot's next state, $s'$: 1) the state label of $s'$ contains all propositions $\pi$ that appear in either $\sigma^T$ or $\sigma^{exT}$ as $\pi^\rho$, and 2) the state label of $s'$ does not contain any propositions $\pi$ that appear in $\sigma^F$ or $\sigma^{exF}$ as $\pi^\rho$. 

A binding assignment $r$ can contain any binding propositions not in $\sigma$, since there are no propositions required by those bindings. In addition, it follows from condition (2) that if a proposition $\pi^\rho$ is in $\sigma^F \cup \sigma^{exF}$ and $\pi$ is not in $AP_j$ (e.g. $camera^1 \in \sigma^F$ and the robot does not have the capability $\lambda_{camera}$), 
$\rho$ can still be assigned to robot $j$. 

 Given the binding assignment function $\mathfrak{R}$, and as shown in \cite{Fang2024}, we can now define the product automaton:
 
 \begin{definition}[Product Automaton] $\mathcal{G}_j = A_j \times \mathcal{B} = (Q, q_0, AP_j,\delta_\mathcal{G},L_\mathcal{G}, W_\mathcal{G}, F_\mathcal{G})$, where 

\begin{itemize}[leftmargin=*]
    \item $Q = S \times Z$ is a finite set of states
    \item $q_0 = (s_0, z_0) \in Q$ is the initial state
    \item $\delta_\mathcal{G}\subseteq Q\times Q$ is the transition relation, where for $q = (s, z)$ and $q' = (s',z')$, $(q,q')\in \delta_\mathcal{G}$ if and only if $(s,s') \in \gamma$ and $\exists \sigma \in \Sigma_{\mathcal{B}}$ such that $(z, \sigma, z') \in \delta_{\mathcal{B}}$ and $\mathfrak{R}(s, \sigma, s') \neq \emptyset$ 
    


    \item $L_\mathcal{G}$ is the labeling function s.t. for $q = (s, z)$, $L_\mathcal{G}(q)\!=\!L(s)\!\subseteq\!AP_j$ 
    \item $W_\mathcal{G}: \delta_{\mathcal{G}} \rightarrow \mathbb{R}_{\ge 0} $ is the cost function s.t. for $(q, q') \in \delta_{\mathcal{G}}$, $q = (s, z)$, $q' = (s',z')$, $W_\mathcal{G}((q, q'))=W((s, s'))$
    \item $F_\mathcal{G} = S \times F$ is the set of accepting states

\end{itemize}

 \end{definition}

\subsection{Adding Intermediate Transitions to the \buchi Automaton} \label{sec:buchi_int}

To ensure correct team behavior when actions take different time durations, we add intermediate transitions to the \buchi
automaton $\mathcal{B}$ to reason about and enforce an ordering on individual robot behaviors (Alg. \ref{algo:buchi_update}). In doing so, we maintain the symbolic abstractions of actions while providing guarantees for the satisfaction of the task by a physical system. 

To motivate the use of intermediate transitions, we look at the transitions $\sigma_{11, i}$ (the inner self-transition along state 1) and $\sigma_{10}$ of the \buchi automaton in Fig \ref{fig:buchi}, where $\sigma_{11, i} = (\sigma_{11, i}^T,\sigExT_{11, i},\sigma_{11, i}^F, \sigExF_{11, i}) = (\emptyset, \emptyset, \{dock_c^1\}, \emptyset)$, $\sigma_{10} = (\sigma_{10}^T,\sigExT_{10},\sigma_{10}^F, \sigExF_{10}) = (\{roomB_c^2, camera^2, roomB_c^3, camera^3, dock_c^1\}, \emptyset, \emptyset, \emptyset)$. $\sigma_{10}$ requires all robots assigned binding 2 and 3 to be in roomB. Since it is unrealistic for multiple robots to enter the room simultaneously, we can leverage the fact that $\sigma_{11, i}$ does not explicitly assign truth values to $roomB_c^2$ and $roomB_c^3$ (i.e. these propositions can have any truth value over $\sigma_{11, i}$) in order to enforce that all the robots assigned bindings 2 and 3 enter room B at some point before executing the transition $\sigma_{10}$. For clarity in the remainder of the paper, we define a proposition $\pi^\rho$ to be assigned an \textit{explicit} truth value over a transition $\sigma$ if it appears in $\sigma$, i.e. $\pi^\rho \in \sigma^T \cup \sigExT \cup \sigma^F \cup \sigExF$. If a proposition is not explicitly in $\sigma$, it can maintain any truth value over that transition.

To automatically construct intermediate transitions, we first define the intermediate propositions function, $\noninstFunc(\sigma)$:

\begin{definition}[Intermediate Propositions Function]
    $\noninstFunc(\sigma): \Sigma_{\mathcal{B}} \rightarrow 2^{AP_{non-inst}} \times 2^{AP_{non-inst}} \times 2^{AP_{non-inst}} \times 2^{AP_{non-inst}}$ such that for $\sigma = (\sigma^T, \sigExT, \sigma^F, \sigExF) \in \Sigma_{\mathcal{B}}$, $\noninstFunc(\sigma)= (I_T, I_{exT}, I_F, I_{exF})$, where $ \forall k \in \{T, exT, F, exF\}, I_k = \{\pi^{\rho} \in \sigma^k \ |  AP_{non-inst} \cap (C_k \in \mathfrak{C}(\sigma, \rho)) \neq \emptyset \}$ 
\end{definition}
$I_T$, $I_{exT}$, $I_F$, and $I_{exF}$ are the sets of \ltlpsi\ propositions corresponding to non-instantaneous actions that are assigned truth values over $\sigma^T, \sigExT, \sigma^F, \text{ and }\sigExF$, respectively. For example, in Fig. \ref{fig:buchi} where $\sigma_{11, o} = (\sigma_{11, o}^T,\sigExT_{11, o},\sigma_{11, o}^F, \sigExF_{11, o}) = (\{roomB_c^2, camera^2, roomB_c^3, camera^3\}, \emptyset, \emptyset, \{dock_c^1\})$, the intermediate propositions are $\noninstFunc(\sigma_{11, o}) = (\{roomB_c^2, roomB_c^3\}, \emptyset, \emptyset, \{dock_c^1\})$, since all the motion propositions are non-instantaneous; $camera$ is instantaneous and therefore omitted from $\noninstFunc(\sigma_{11, o})$.

We use the intermediate propositions function to determine which non-instantaneous propositions change truth values across two transitions; we then use these propositions to enforce constraints on the binding assignments and on robot behavior. Let $\sigma, \sigma'$ be labels of two transitions in the \buchi automaton, and $\noninstFunc(\sigma) = (I_T, I_{exT}, I_F, I_{exF}), \noninstFunc(\sigma') = (I'_T, I'_{exT}, I'_F, I'_{exF})$.  To categorize how these propositions change their truth values, we define the following sets (Alg. \ref{algo:buchi_update}, line \ref{algo:B_upd:Pisets}): 
\begin{align}
    \PiFT &= I_T \cap (I'_F \cup I'_{exF}) \\
    \PiTF &= I_F \cap (I'_T \cup I'_{exT} )
\end{align}

$\PiFT, \PiTF \subseteq AP_{non-inst}$ are the sets of non-instantaneous \textit{for all} propositions that explicitly change truth values between $\sigma$ and $\sigma'$. 
\begin{align}
    \PiexT &= I_{exT} \cap (I'_F \cup I'_{exF}) \\
    \PiexF &= I_{exF} \cap (I'_T \cup I'_{exT})
\end{align}
$\PiexT, \PiexF \subseteq AP_{non-inst}$ include any non-instantaneous proposition that is a \textit{there exists} proposition in $\sigma$ that explicitly changes truth values between $\sigma$ and  $\sigma'$. 
\begin{align}
    \PixT &= (I'_T \setminus I_T) \\
    \PixexT &=(I'_{exT} \setminus I_{exT}) \\
    \PixF &= (I'_F\ \setminus I_F) \\
    \PixexF &= (I'_{exF} \setminus I_{exF})
\end{align}

$\PixT, \PixexT, \PixF, \PixexF \subseteq AP_{non-inst}$ are the sets of non-instantaneous propositions that are assigned truth values in $\sigma'$ but not in $\sigma$. 

\textit{Example: Task 1.} In Fig. \ref{fig:buchi}, consider the two transitions $(1, \sigma_{11, i}, 1)$, $(1, \sigma_{10}, 0)$, where $\sigma_{11, i} = (\emptyset, \emptyset, \{dock_c^1\}, \emptyset)$ is the inner self-transition on state 1, and $\sigma_{10} = (\{roomB_c^2, camera^2, roomB_c^3, camera^3, dock_c^1\}, \emptyset, \emptyset, \emptyset)$. Then $\PiTF = \{dock_c^1\}$, $\PixT = \{roomB_c^2, roomB_c^3\}$; the remaining sets are empty.

\begin{algorithm}
    \SetKwInOut{Input}{Input}
    \SetKwInOut{Output}{Output}
    \SetKwProg{Initialization}{Initialization}{}{}
    \Input{$\mathcal{B}, e_1 = (z_1, \sigma_{12}, z_2), e_2 = (z_2, \sigma_{23}, z_3)$
    , where $\sigma_{12} = (\sigma_{12}^{T}, \sigma_{12}^{exT},\sigma_{12}^{F},\sigma_{12}^{exF})$, $\sigma_{23} = (\sigma_{23}^T, \sigma_{23}^{exT},\sigma_{23}^{F},\sigma_{23}^{exF})$} 
    \Output{$\buchiUpd, E_{upd}, (c_{single}, c_{combo})$}

    $\mathcal{B}_{upd} = \mathcal{B}$,
    $c_{single} = \emptyset, c_{combo} = \emptyset$ \\
    $(I_T, I_{exT}, I_F, I_{exF}) = \noninstFunc(\sigma_{12}),  (I'_T, I'_{exT}, I'_F, I'_{exF}) = \noninstFunc(\sigma_{23})$ \\
    \tcp{compute sets of non-inst actions}
    $\PiFT, \PiTF, \PiexT, \PiexF, \PixT, \PixF, \PixexT, \PixexF = \textsc{non\_inst\_sets} ((I_T, I_{exT}, I_F, I_{exF}), (I'_T, I'_{exT}, I'_F, I'_{exF}))$ \label{algo:B_upd:Pisets} \\
    \tcp{keep track of binding constraints}
    \If{$|\PiFT \cup \PiTF| \geq 1$ \label{B_upd:binding_constraints1}}{
        $c_{single} = \mathfrak{B}_\Pi(\PiFT \cup \PiTF)$}
    \If{$|\PiexT \cup \PiexF| \geq 2$}{
        $c_{combo} = \mathfrak{B}_\Pi(\PiexT \cup \PiexF)$ \label{B_upd:binding_constraints2}
    }
    \tcp{no need to update \buchi}
    \If{$|I_T' \cup I_{exT}' \cup I_F' \cup I_{exF}'| \leq 1$ or $|\PiFT \cup \PiTF \cup \PiexT \cup \PiexF| < 1$ \label{B_upd:noupdate}}{
    \Return $\mathcal{B}, \emptyset, c_{single}, c_{combo}$
    }    
    $\sigIntOne = (\sigIntOne^{T}, \sigIntOne^{exT}, \sigIntOne^{F}, \sigIntOne^{exF}) = \textsc{intermediate\_trans}(\sigma_{12}, \PixT, \PixexT, \PixF, \PixexF)$ \label{B_upd:sig_int}
    \If{$z_1 = z_2$ \label{B_upd:self-nonself1}}{
        add $(z_1, \sigma_{12}, z_{1^{*}})$, $(z_{1^{*}}, \sigIntOne, z_{1^{*}})$,  $(z_{1^{*}}, \sigma_{23}, z_{3})$ to $\mathcal{B}_{upd}$; remove $(z_2, \sigma_{23}, z_3)$ from $\mathcal{B}_{upd}$\\
        $E_{upd} = ((z_1, \sigma_{12}, z_{1^{*}})$, $(z_{1^{*}}, \sigIntOne, z_{1^{*}})$,  $(z_{1^{*}}, \sigma_{23}, z_{3}))$} \label{B_upd:self-nonself2}
    \Else{     
        \tcp{modify first transition (non-self transition)} 
        
        \If{$\sigma_{12} \neq \sigIntOne$}{
        modify $(z_1, \sigma_{12}, z_2) = (z_1, \sigIntOne, z_2)$ in $\mathcal{B}_{upd}$ \\
        $E_{upd} = ((z_1, \sigIntOne, z_2))$}
        \Else{$E_{upd} = ()$}
        }

    \Return $\buchiUpd, E_{upd}, (c_{single}, c_{combo})$
\caption{Update \buchi Automaton}
\label{algo:buchi_update}
\end{algorithm}

Violations of the task may occur when the robots execute a transition in the \buchi automaton $\mathcal{B}$. Thus, to prevent any violations, we check two sequential transitions in $\mathcal{B}$, $e_1 = (z_1, \sigma_{12}, z_2), e_2 = (z_2, \sigma_{23}, z_3)$, and update the \buchi automaton $\mathcal{B}$ if a specific ordering of propositions is necessary. {When there is at least one non-instantaneous proposition that changes truth values along $e_2$, we want to enforce an ordering such that the other non-instantaneous propositions are completed first. Thus, we only need to modify the \buchi automaton if there exists more than one non-instantaneous proposition appears along $e_2$ ($|I_T' \cup I_{exT}' \cup I_F' \cup I_{exF}'| > 1$) and at least one of them explicitly changes truth values ($|\PiFT \cup \PiexT \cup \PiTF \cup \PiexF| \geq 1$, line \ref{B_upd:noupdate})}. 

\textbf{Binding Constraints (lines \ref{B_upd:binding_constraints1} - \ref{B_upd:binding_constraints2}): }
There may be a sequence of transitions in which non-instantaneous propositions change truth values, which may require imposing constraints of the robot binding assignments. 
For example, looking at the same two transitions as before, $(1, \sigma_{11, i}, 1)$ and $(1, \sigma_{10}, 0)$, $\PiFT \cup \PiTF = \{dock_c^1\}$. In $\sigma_{11, i}$, we require all robots assigned binding 1 to not be in the dock, and in the next transition $\sigma_{10}$, we require all robots assigned binding 1 to be in the dock. If multiple robots are assigned binding 1, this realistically will not happen when executing in continuous time, as it would require all robots to enter the dock at the exact same time. Thus, to guarantee these transitions in the \buchi automaton are satisfied, we constrain the binding assignments such that one and only one robot can be assigned bindings in $\mathfrak{B}(\PiFT \cup \PiTF)$ (in this case, binding 1). The same constraints apply if multiple bindings are present (e.g. if $\PiFT \cup \PiTF = \{dock_c^1, roomB_c^2\}$, then one and only one robot can be assigned both bindings 1 and 2; otherwise, two robots must make the transition at the exact same time, which is not realistic). These combinations of bindings are stored in $c_{single}$.

Similar constraints apply when $|\PiexT \cup \PiexF| \geq 2$, i.e. multiple \textit{there exists} propositions change truth values. In this case, for each proposition $\pi^\rho$ in $\PiexT \cup \PiexF$, we require that at least one robot assigned $\rho$ will change its truth value of $\pi_c$ across the two transitions. For instance, if $\sigma_{12} = \{ \emptyset, \emptyset, \emptyset, \{dock_c^1, push_c^2\}\}$, $\sigma_{23} = \{\{dock_c^1, push_c^2\}, \emptyset, \emptyset, \emptyset\}$), $\sigma_{12}$ requires at least one robot assigned binding 1 to not be at the dock, and at least one robot assigned binding 2 to not complete the pushing action, while $\sigma_{23}$ requires all robots assigned binding 1 and 2 to be at the dock and complete pushing, respectively. On a physical system, it is unrealistic for robots to be able to synchronize completing a non-instantaneous action at the same time. As a result, to guarantee satisfiability, we constrain the binding assignment such that if a robot is assigned to any binding in $\mathfrak{B}_\Pi(\PiexT \cup \PiexF)$, then it must be assigned to all bindings within that set; otherwise, it cannot be assigned to any of them.
These binding assignments are stored in $c_{combo}$. 
Note that the constraints are not necessary when $|\PiexT \cup \PiexF| = 1$; in this case, only one robot assigned $\rho$ must change its truth value across the two transitions, which can easily be satisfied by a physical system.

If more than one non-instantaneous proposition appears along the second transition $e_2$, we generate an intermediate transition $\sigIntOne = (\sigIntOne^{T}, \sigIntOne^{exT}, \sigIntOne^{F}, \sigIntOne^{exF})$ (line \ref{B_upd:sig_int}), where $\sigIntOne^{T} = \sigma_{12}^{T} \cup \Pi_{\emptyset T}$, 
$\sigIntOne^{F} = \sigma_{12}^{F} \cup \Pi_{\emptyset F}$ , $\sigIntOne^{exT} = \sigma_{12}^{exT} \cup \Pi_{\emptyset exT}$, 
$\sigIntOne^{exF} = \sigma_{12}^{exF} \cup \Pi_{\emptyset exF}$; because any proposition that does not appear in $\sigma_{12}$ can maintain any truth value, the transition $\sigIntOne$ ensures that all propositions that appear in $\sigma_{23}$ but not in $\sigma_{12}$ (i.e. $\PixT \cup \PixexT \cup \PixF \cup \PixexF$) change their truth values, in any order, before any of the propositions that change truth values between the transitions (i.e. $\PiTF \cup \PiexT \cup \PiFT \cup \PiexF$) do so.



\textbf{Updating transitions in $\mathcal{B}$: } Since we assume that every state in the \buchi automaton $\mathcal{B}$ has a self transition, either $e_1$ is a self-transition and $e_2$ is not, or vice-versa; the approach in updating $\mathcal{B}$ differs for the two cases. 

\begin{figure*}
    \centering
        \includegraphics[width=0.95\textwidth]{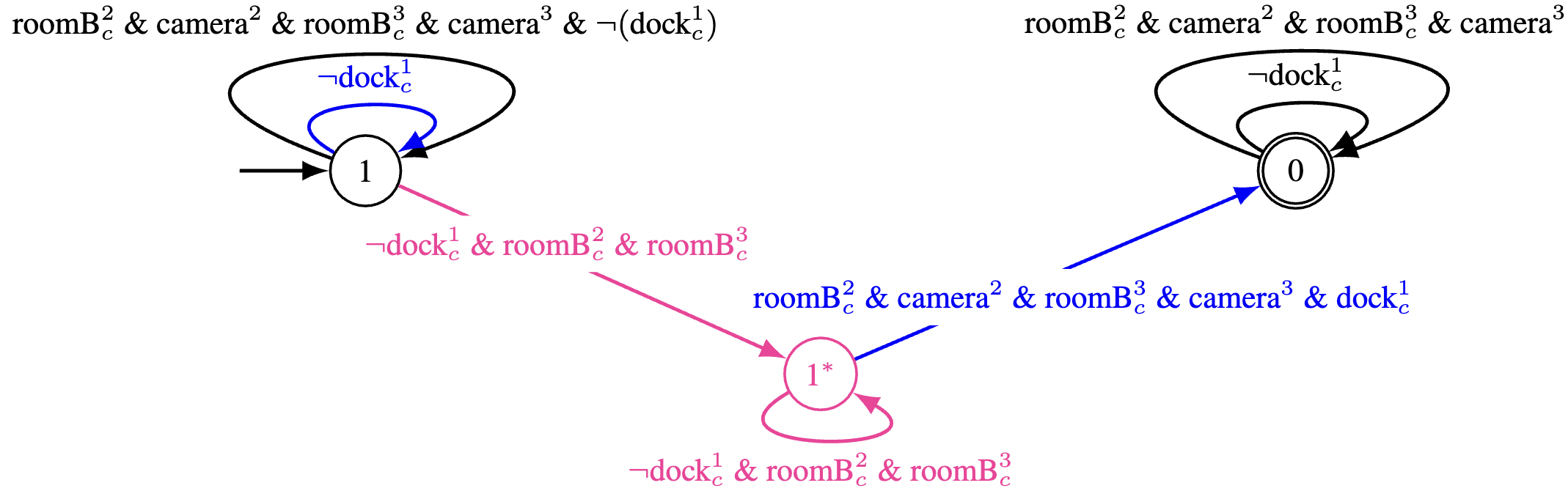}
        \label{fig:buchi_int1}
    \caption{Updated \buchi automaton for Case 1 ($e_1$ is a self-transition and $e_2$ is not). The blue represents the original transitions, the pink represents the added intermediate states and transitions}
    \label{fig:buchi_int}
\end{figure*}

\textbf{Case 1:} \textit{$e_1$ is a self-transition and $e_2$ is not (lines \ref{B_upd:self-nonself1} - \ref{B_upd:self-nonself2}).} We take advantage of the fact that $e_1$ is a self-transition and add an intermediate state along with the transitions $(z_1, \sigma_{12}, z_{1^{*}})$, $(z_{1^{*}}, \sigIntOne, z_{1^{*}})$, and $(z_{1^{*}}, \sigma_{23}, z_3)$. These transitions satisfy the self-transition $e_1$ while also including constraints that must be enforced to ensure the robot's continuous time execution of the actions do not violate any part of $e_2$.

\textit{Example:} Given the \buchi automaton in Fig. \ref{fig:buchi}, let $e_1 = (1, \sigma_{11, i}, 1)$, $e_2 = (1, \sigma_{10}, 0)$, where $\sigma_{11, i}$ and $\sigma_{10}$ are defined earlier in the example. Then $\sigIntOne = (\{roomB_c^2, roomB_c^3\}, \emptyset, \{dock_c^1\}, \emptyset)$. We add an intermediate state $1^*$ and the corresponding transitions $(1, \sigma_{11, i}, 1^*)$, $(1^*, \sigIntOne, 1^*)$, and $(1^*, \sigma_{10}, 0)$. Fig. \ref{fig:buchi_int} shows the \buchi automaton with these updates.

\textbf{Case 2:} \textit{$e_2$ is a self-transition and $e_1$ is not.} Because $e_1$ is not a self-transition, we cannot add intermediate states between $e_1$ and $e_2$. In this case, we modify $e_1$ from $(z_1, \sigma_{12}, z_2)$ to $(z_1, \sigIntOne, z_2)$. Intuitively, we are adding all the non-instantaneous propositions explicitly in $\sigma_{23}$ into $\sigma_{12}$. {Changing this transition does not affect the satisfiability of the original task, since the modification only adds more constraints on propositions that do not violate the original formula along the transition.
} 

\textit{Example:} Given the \buchi automaton in Fig. \ref{fig:buchi}, let $e_1 = (1, \sigma_{10}, 0)$, $e_2 = (0, \sigma_{00}, 0)$ , where $\sigma_{10}$ is defined earlier in the example, and $\sigma_{00} = (\emptyset, \emptyset, \{dock_c^1\}, \emptyset)$. Since both $\PixT$ and $\PixF$ are empty, $\sigIntOne$ and $\sigma_{10}$ are equivalent, and no updates to the \buchi automaton are necessary. However, since $\PiFT \cup \PiTF = \{dock_c^1\}$, one and only one robot can be assigned binding 1. Thus, $c_{single} = \{1\}$.

Algorithm \ref{algo:buchi_update} outputs $\mathcal{B}_{upd}$, the updated \buchi automaton, $E_{upd}$, the edges that have been modified/added, and $(c_{single}, c_{combo})$, which include constraints on the bindings in which one and only one robot can be assigned to (i.e. $c_{single}$), or bindings that a robot must be assigned either to all or none of those bindings (i.e. $c_{combo}$).



\subsection{Updating the Product Automaton} \label{sec:update_prod_aut}

\begin{algorithm}
    \SetKwInOut{Input}{Input}
    \SetKwInOut{Output}{Output}
    \SetKwProg{Initialization}{Initialization}{}{}
    \Input{$\mathcal{B}, \buchiUpd, E_{upd}, \mathcal{G}_j$} 
    \Output{$\prodUpd$}

    \If{$\buchiUpd = \mathcal{B}$}{
        \Return $\mathcal{G}_j$
    }
    $\prodUpd = \mathcal{G}_j$ \\
        
        
    \If {$|E_{upd}| > 1$}{    
        $(z_1, \sigma_{12}, z_{1^{*}}),(z_{1^{*}}, \sigIntOne, z_{1^{*}}), (z_{1^{*}}, \sigma_{23}, z_{3}) = E_{upd}$ \\
        $\prodUpd = \textsc{remove\_edges}(\prodUpd, (z_2, \sigma_{23}, z_3))$ \\
        $\buchiUpd' = \textsc{get\_subBuchi}(\buchiUpd, E_{upd})$ \\
        $A_j' = \textsc{get\_subAgent}(\mathcal{G}_j, (z_1, \sigma_{12}, z_2), (z_2, \sigma_{23}, z_3))$ \\
        $\mathcal{G}_j' = A_j' \times \buchiUpd'$ \\
        
        \tcp{join two product automata} 
        $\prodUpd = \prodUpd \cup \mathcal{G}_j'$
    }
    \ElseIf{$|E_{upd}| = 1$}{
        $(z_1, \sigIntOne, z_2) = E_{upd}$ \\
        \tcp{sub product automaton to check} 
        $\mathcal{G}_j' = \textsc{get\_subProd}(\mathcal{G}_j, (z_1, \sigma_{12}, z_2))$

        \For{$(q, q') \in \delta_{\mathcal{G}_j'}$}{
            \If{not $\textsc{valid\_edge}((q, q'), \sigIntOne)$}{
            $\prodUpd.remove\_edge(q, q')$
            
            }
        }
    }
    \Return $\prodUpd$
    
\caption{Update Robot $j$'s Product Automaton}
\label{algo:prodaut_update}
\end{algorithm}

Because we are modifying the \buchi automaton as we search for an accepting trace through it, we also need to modify each robot's product automaton $\mathcal{G}$ to check if the robot can synthesize controllers to satisfy the given trace in the \buchi automaton (Alg. \ref{algo:prodaut_update}). {This is done for each transition we check in the \buchi automaton, when constructing the trace}.  

As mentioned in Sec. \ref{sec:buchi_int}, there are two types of modifications to the \buchi automaton depending on whether the first transition is a self-transition and the second is not, or vice-versa; as a result, there are two ways to update the product automaton. In both cases, we take advantage of the fact that $\sigma_{12} \subseteq \sigIntOne$ (i.e. the propositions in $\sigIntOne$ may have constraints on the truth values of propositions in addition to the ones in $\sigma_{12}$); rather than reconstructing robot $j$'s product automaton $\mathcal{G}_j$ by taking the entire cross product $A_j \times \buchiUpd$, we modify portions of the existing $\mathcal{G}_j$ based on the changes made to $\mathcal{B}$ to get $\mathcal{G}^{upd}_j$.

Let the two transitions we checked in the original \buchi automaton $\mathcal{B}$ be $e_1 = (z_1, \sigma_{12}, z_2), e_2 = (z_2, \sigma_{23}, z_3)$, where $\sigma_{12} = (\sigma_{12}^{T}, \sigma_{12}^{exT},\sigma_{12}^{F},\sigma_{12}^{exF})$, $\sigma_{23} = (\sigma_{23}^T, \sigma_{23}^{exT},\sigma_{23}^{F},\sigma_{23}^{exF})$, and the set of updated transitions $E_{upd}$ is outputted by Alg. \ref{algo:buchi_update}. 

\textbf{Case 1:} \textit{$e_1$ is a self-transition and $e_2$ is not (i.e. $|E_{upd}| > 1$).} 
In this case, the transition $e_2$ has been removed and replaced with intermediate states and associated transitions in the \buchi automaton (Alg. \ref{algo:buchi_update}).

To update the product automaton, we first remove any transitions of the product automaton $((s,z_2), (s',z_3)) \in \delta_{\mathcal{G}_j}$, since we have also removed $e_2$ from $\buchiUpd$. Then, we take the cross product only of the modified portions of each graph and add it to $\prodUpd$.

\textbf{Case 2:} \textit{$e_2$ is a self-transition and $e_1$ is not (i.e. $|E_{upd}| = 1$).} In this case, $\buchiUpd$ does not have any new transitions, but $\sigma_{12}$ has been modified to become $\sigIntOne$. Thus, we check any transition in $((s,z_1), (s',z_2)) \in \delta_{\mathcal{G}_j}$ to see if it is still valid based on $(z_1, \sigIntOne, z_2)$, the updated transition in $\buchiUpd$.

\subsection{Robot Team Behavior} \label{sec:DFS}

To find a team of robots that can satisfy the task, we need to find a trace in the \buchi automaton such that every binding is assigned, with each robot maintaining a consistent set of bindings throughout the entire trace. 

In our prior work, we introduced a DFS algorithm to find a trace $\beta$ in the \buchi automaton, $\beta_{\hat{A}}$, a team of robots $\hat{A}$, and corresponding binding assignments $R_{\hat{A}}$ to collectively execute discrete actions. We extend this framework to continuous actions so that the robots' behaviors are guaranteed to satisfy the original specifications (Alg. \ref{algo:dfs_upd}). 
To do so, during the DFS, for each transition $(z, \sigma, z')$ we check, we update the \buchi automaton if intermediate states are necessary (line \ref{algo:dfs_upd:update_buchi}). Each robot updates its product automaton, and determines which bindings it can satisfy (line \ref{algo:dfs_upd:update_prodaut}).

We want the robot's behavior to satisfy not only the current transition, but also the entire path with a fixed binding assignment. To ensure this, each robot's assigned binding set $r_j$ is first initialized to be the set of all possible binding assignments, and each robot removes binding assignments that are not possible with each iteration of the DFS algorithm (line \ref{line:update_combos1}). It does so by finding a trace through its product automaton following the current trace in the \buchi automaton, $\beta_{\hat{A}} \cup E_{upd}$. The robots also take the binding constraints $(c_{single}, c_{combo})$ into account, reducing the number of combinations of bindings the robots need to check (e.g. if $c_{combo} = \{1,2\}$
, then a robot assigned binding 1 must also be assigned binding 2; there is no need for the robot to check its ability to satisfy the bindings separately). If $r_j = \emptyset$ (i.e. the robot cannot be assigned any binding), we remove the robot from the team $\hat{A}$ entirely.

$c_{single}$ is the set of bindings assignments for which exactly one robot must be assigned, but
the framework may output a teaming assignment that includes multiple robots assigned those bindings. We keep track of these binding constraints and ask the user to choose one robot for each $c \in c_{single}$. 

Because we are using a DFS approach to find a viable team, the final teaming assignment may not be the globally optimal one. However, the framework is sound; it is guaranteed to find an assignment if one exists.

\begin{algorithm}[t]
    \SetKwInOut{Input}{Input}
    \SetKwInOut{Output}{Output}
    \SetKwProg{Initialization}{Initialization}{}{}
    \Input{$A=\{A_1, A_2, ..., A_n\}$, $\mathcal{B}$, \new{$ G = \{ \mathcal{G}_1, \mathcal{G}_2 ..., \mathcal{G}_n \}, \APbinding$}}
    \Output{$\beta$, $\delta_{self}$, $\hat{A} \subseteq A$, $R_{\hat{A}}$, \new{$G_{\hat{A}}, \buchiUpd$}}
    
    $stack = \emptyset$, $visited = \emptyset$  \\
    \new{$R = \{r_k = 2^{\APbinding} \setminus \emptyset \ | \ k = \{1,2,...,n\} \}$} \\
    \For{$e \in \{(z, \sigma, z') \in \delta_{\mathcal{B}} \ | \ z = z_0 \}$ }{
        $stack = stack \cup \{(e, R, [e], \new{\mathcal{B}, G, \emptyset})\}$ 
    }

    \While{$stack \neq \emptyset$}{
        $((z, \sigma, z'), R_{\hat{A}}, \beta_{\hat{A}}, \new{\mathcal{B}, G_{\hat{A}}, C}) = stack.pop()$ \\
        
        \If{$(z, \sigma, z') \not\in visited$}{
        $visited = visited \cup (z, \sigma, z')$ \\
        \new{
        \tcp{update buchi automaton with intermediate states}
        \If{$|\beta_{\hat{A}}| > 1$}{
        $e_1 = \beta_{\hat{A}}[-1], $
        $e_2 = (z, \sigma, z')$ \\
        $\buchiUpd, E_{upd}, (c_{all}, c_{ex}) = \textsc{update\_buchi}(\mathcal{B}, e_1, e_2)$ \label{algo:dfs_upd:update_buchi}
        }
        \Else{
        $\buchiUpd, E_{upd}, (c_{all}, c_{ex}) = \mathcal{B}, \emptyset, (\emptyset, \emptyset)$
        }
        \For{$\mathcal{G}_j \in G_{\hat{A}}$}{
        $\prodUpd = \textsc{update\_product\_aut}(\buchiUpd, E_{upd}, \mathcal{G}_j)$
        \label{algo:dfs_upd:update_prodaut}
        }
        }

        
        \For{$r_j \in R_{\hat{A}}$ 
        }{ \label{line:update_combos1}
            $r_j' = \textsc{update\_bindings}(r_j$, $\new{\prodUpd, \beta_{\hat{A}} \cup E_{upd}, (c_{all}, c_{ex})})$  \\
            \If{$r_j' = \emptyset$}{
                $R_{\hat{A}} = R_{\hat{A}} \setminus r_j$, $\new{G_{\hat{A}} = G_{\hat{A}} \setminus \mathcal{G}_j}$
            }
            \Else{
            $R_{\hat{A}} = (R_{\hat{A}} \setminus r_j) \cup r_j'$}

        }
        \label{line:update_combos2}

        \If{$\bigcup_j (R_j \in R_{\hat{A}}) = \APbinding$}{
            
            \If{$z' \in F$}{ 
            $\beta, \delta_{self} = \textsc{parse\_trace}(\beta_{\hat{A}})$ \\
            \Return $\beta, \delta_{self}, R_{\hat{A}}, \new{\buchiUpd, C \cup \{(c_{all}, c_{ex})\}}$}
            
            
            $E = \{(z',\sigma', z'') \in \delta_{\mathcal{B}}\}$ \label{line:E_z}
            
            \For{$(z',\sigma', z'') \in  E$
                }{
                \If{$(z = z'  \text{ and } z'\neq z'')$ or 
                 $(z \neq z'  \text{ and } z'= z'')$ \label{line:check_self}}{
                    $stack = stack \cup \{ (z',\sigma', z''), R_{\hat{A}}$,  \new{$\beta_{\hat{A}} \cup E_{upd}, \buchiUpd, G_{\hat{A}}, C \cup \{(c_{all}, c_{ex})\}$}
                    }
                }
            
            }
        
        }
    }
\caption{Find Accepting Trace for Robot Team}
\label{algo:dfs_upd}
\end{algorithm}

\subsection{Synthesis and Execution of Control and Synchronization Policies} \label{sec:sync}
To guarantee that the behavior of the robots do not violate the original task, the robots must implement synchronization policies to transition to the next non-intermediate state in $\mathcal{B}_{upd}$ at the same time. The algorithm (Alg. \ref{algo:synthesis}) is similar to \cite{Fang2024}; a robot $j$ first synthesizes its behavior $b_j$ for the entirety of the task that satisfies the collective trace in the \buchi automaton $\beta$ (function $\textsc{find\_behavior}$, line \ref{algo:sync:b_j}). For each transition $(z, \sigma, z')$ in $\beta$, the robot parses out $b_j^{zz'}$, the behavior that correlates to $(z, \sigma, z')$ (function $\textsc{sub\_behavior}$, line \ref{algo:sync:sub_b_j}).  
Before it executes this behavior, it checks if it is required to participate in the synchronization step. It participates if 1) $z'$ is not an intermediate state, 2) $r_j$ contains a binding $\rho$ required by $\sigma$ and 3) is not the only robot assigned bindings from $\sigma$ (line \ref{line:Rbar}). If the robot satisfied these criteria, its binding assignment $r_j$ is added to $\overline{R}$, which contains the binding assignments of every robot that needs to participate in the synchronization step at state $z'$.


One difference of this algorithm compared to our previous approach \cite{Fang2024} is that we do not need to execute the synchronization policy if $z'$ is an intermediate state (Alg. \ref{algo:synthesis}.
lines (\ref{line:new_int} - \ref{line:new_int_end})), since the synchronization is only required on the transitions of the original \buchi automaton.

The other key difference is in lines \ref{alg:synth:extra_comm1} - \ref{alg:synth:extra_comm2} of Alg. \ref{algo:synthesis}. Because instantaneous actions may need to synchronize when another action finishes (e.g. a robot must take a camera as soon as another robot enters room B), robots also need to communicate when all of their non-instantaneous actions finish executing (e.g. when $roomBc$ is true) by executing the function $\textsc{execute\_noninst}$ (line \ref{alg:synth:extra_comm1}). As soon as all the non-instantaneous actions across the robots in $\overline{R}$ are complete, the robots execute their instantaneous actions (function $\textsc{execute\_inst}$, line \ref{alg:synth:extra_comm2}).

The waiting state for each robot $z_{wait}$ is the penultimate state in its behavior, $b_j[\ell-1]$. Each robot stays in $z_{wait}$ until all the other robots in $\overline{R}$ are also in their waiting states. Then, the robots execute the last 
step in their behavior, $b_j[\ell]$, synchronously.

\begin{algorithm}[t]
    \SetKwInOut{Input}{Input}
    \SetKwInOut{Output}{Output}
    \SetKwProg{Initialization}{Initialization}{}{}
    \Input{$\mathcal{G}_j$, $r_j$, $R_{\hat{A}}$, $\beta$, $\delta_{self}$}
    
    $b_j = \textsc{find\_behavior}(\mathcal{G}_j, r_j, \beta, \delta_{self})$ \label{algo:sync:b_j}

    \For{$(z,\sigma, z') \in \beta$}{
    $b_{j}^{zz'} = \textsc{sub\_behavior}(b_j, (z,\sigma, z'))$ \label{algo:sync:sub_b_j} \\
    \new{
    \label{line:new_int}
    \If{$z'=z^*$}{
        $p = ()$ \\
        $\textsc{execute}(b_{j}^{zz'}, p)$ \\
        \Continue
        \label{line:new_int_end}
    }}
    
    $\overline{R} = \{r_k \in R_{\hat{A}} \ | \ r_k \cap \mathfrak{B}(\sigma) \neq \emptyset \}$ \label{line:Rbar} \\    
    \If{$r_j \not \in \overline{R}$ or $\overline{R}=\{r_j\}$ \label{line:nosync}}{
    $p = ()$ \\
    $\textsc{execute}(b_j, p)$
    }
    \Else{
    $p = (j, z', 0)$, $\ell = length(b_j)$ \\
    $\textsc{execute}(b_{j}^{zz'}[1:\ell-1], p)$
    \label{line:execute}\\
    $z_{wait} = b_j[\ell-1]$, $P = \{j\}$  \\
    \While{$\bigcup_{i \in P} (r_i \in \overline{R}) \neq \mathfrak{B}(\sigma)$}{
       $p = (j, z', 1)$ \\
        $\textsc{execute}(z_{wait}, p)$ \\
        $P = j \cup \{ k \ | \ (k, z', 1) \in \textsc{receive}()\}$ \label{line:check_sync}   
    }
    \new{
    $\Phi = (j,z',0)$ \label{alg:synth:extra_comm1}\\
    $\textsc{execute\_noninst}(b_{j}^{zz'}[\ell], \Phi)$  \\
    $\textsc{execute\_inst}(b_{j}^{zz'}[\ell], ())$ }\label{alg:synth:extra_comm2}
    }
    }     
\caption{Synthesize Robot $j$'s Behavior}
\label{algo:synthesis}
\end{algorithm}

\section{Results}

\subsection{Demonstrations}
\begin{figure}[h!]
    \centering 
    \includegraphics[width=0.9\columnwidth]{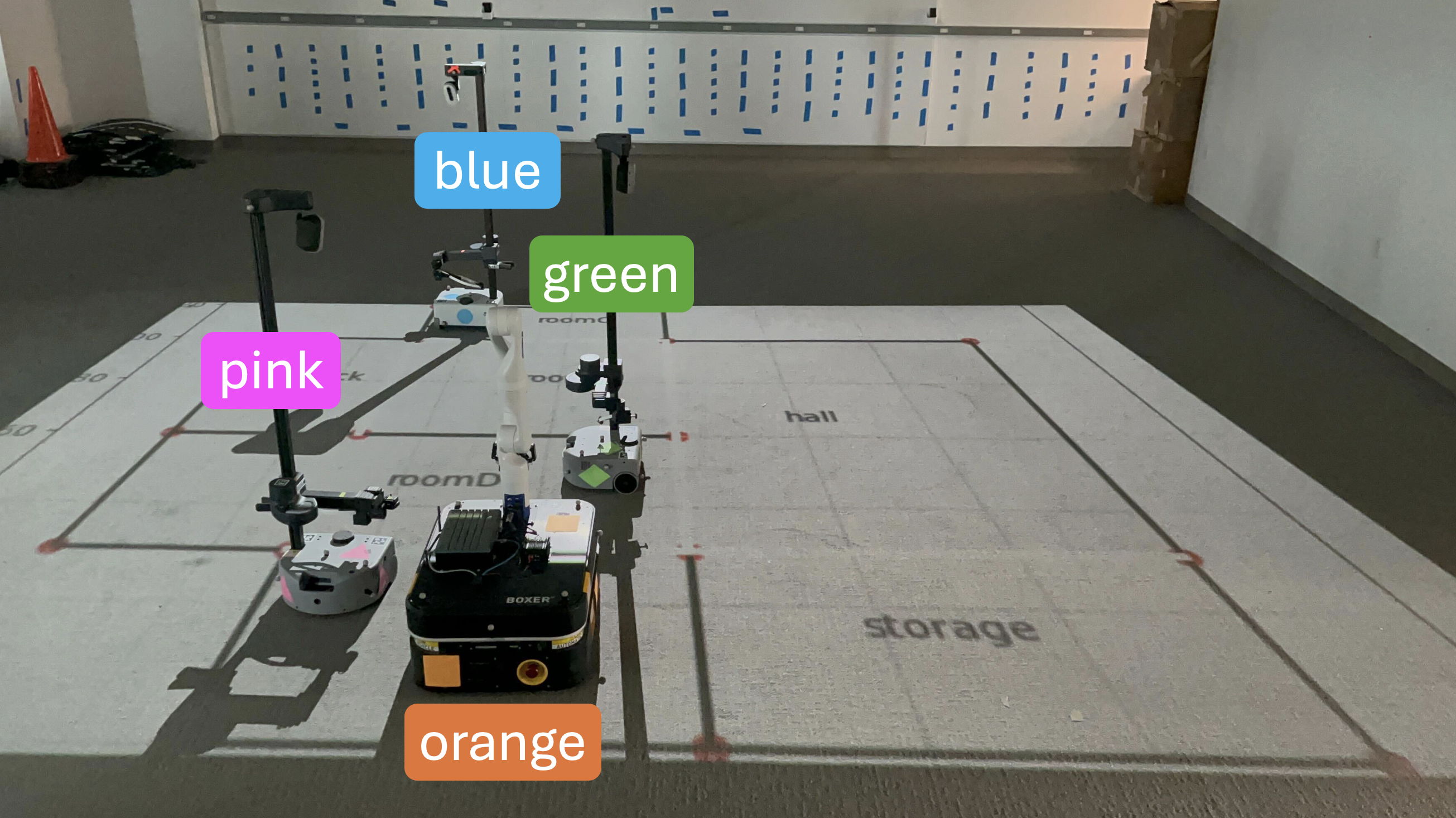}
    \caption{Initial setup on the physical system}
    \label{fig:setup_phys}
\end{figure}

We demonstrate our framework through two different collaborative high-level tasks, executed on physical hardware. We use 4 robots - two Hello Robot Stretch RE1, one  Hello Robot Stretch 2, and a Clearpath Boxer with a Kinova Gen3 arm. We use an Optitrack motion capture system to track the robots and the environment. The setup is shown in Fig. \ref{fig:setup_phys}. 
{Their capabilities are the same as those listed in Sec. \ref{sec:setup}}:

\noindent
$\Lambda_{green} = \{\lambda_{\mathit{mot}}, \lambda_{arm}, \lambda_{camera}\}, L(s_0) = \{roomD_c\}$ \\
$\Lambda_{blue} = \{\lambda_{\mathit{mot}}, \lambda_{beep}\}, L(s_0) = \{roomC_c\}$ \\
$\Lambda_{orange} = \{\lambda_{\mathit{mot}}, \lambda_{arm}\},L(s_0) = \{roomE_c\}$ \\
$\Lambda_{pink} = \{\lambda_{\mathit{mot}},  \lambda_{beep},\lambda_{cam}, \lambda_{scan}\}$, $L(s_0) = \{roomE_c\}$

\noindent{where the green and pink robots are Hello Robot Stretch RE1s, the blue robot is a Hello Robot Stretch 2,  and the orange robot is the Clearpath Boxer with a Kinova Gen3 arm. Although all the Hello Robot Stretch robots have arms, we treat the blue and pink robots as not having those capabilities.} The demonstrations of the full behavior is shown in the accompanying video\footnote{\url{https://youtu.be/DiWjqXHQjmI}}.

\subsection{Example: Task 1}

\begin{figure}[h!]
    \centering 
    \includegraphics[width=0.71\columnwidth]{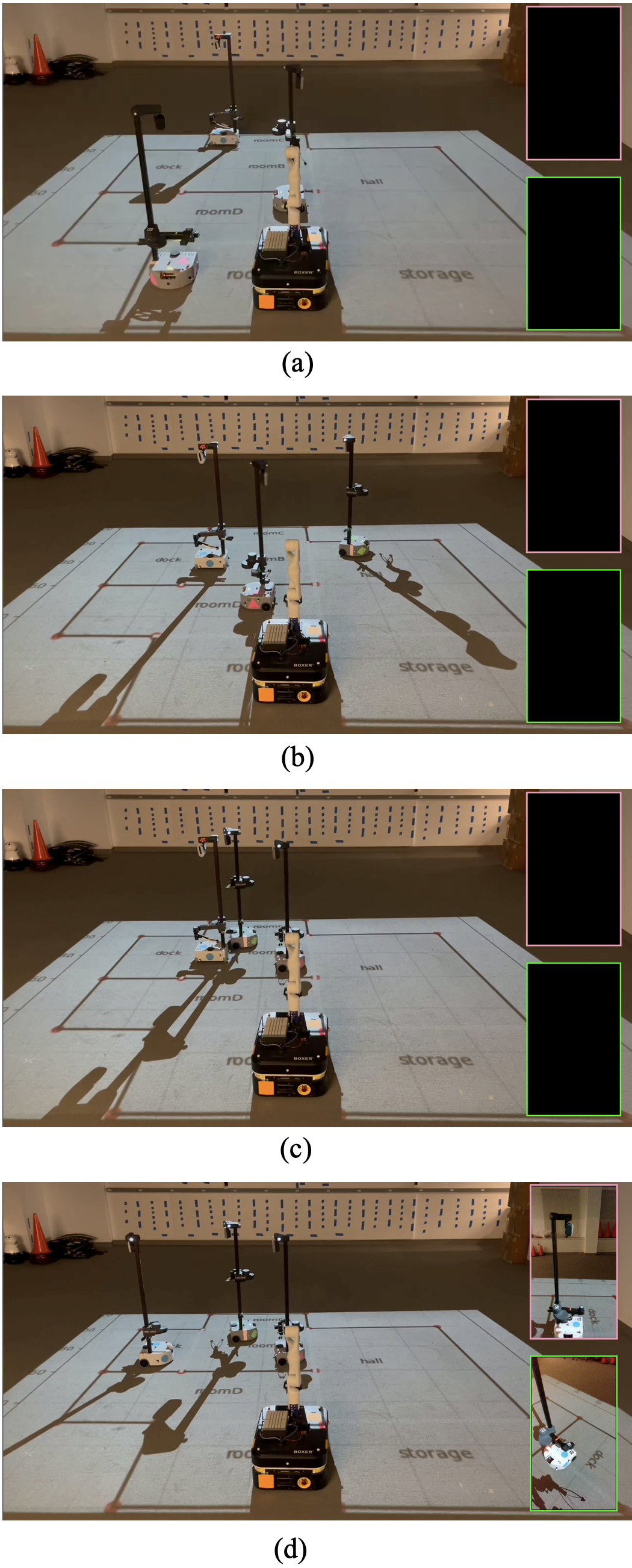}
    \caption{Robots executing task 1. The boxes on the right in each frame represents the pink and green robots' cameras.}
    \label{fig:ex1}
\end{figure}

We implement our framework for the team of robots to satisfy the task defined in Eq. \ref{ex:ex1}. 
The team created by the framework is $\hat{A}=\{$$A_{green}$, $A_{blue}$, $A_{orange}$, $A_{pink}\}$ with binding assignments $r_{green} = \{2,3\}$, $r_{blue} = \{1\}, r_{orange} = \{1\}, r_{pink} = \{2,3\}$ with the constraints $c_{all} = \{\{1\}\}$. In this case, exactly  one robot can be assigned binding 1, so the user chooses between $A_{blue}$ and $A_{orange}$ to be part of the team. If we choose between the two based on minimizing cost, the final team is $\hat{A}=\{$$A_{green}$, $A_{pink}$, $A_{blue}\}$.

Fig. \ref{fig:ex1} provides key frames in the robots' execution of the task. The boxes on the right in each frame represents the pink and green robots' cameras. We can see that the blue robot waits outside the dock area while the pink and green robots make their way to roomB (Fig. \ref{fig:ex1}b). After they are in roomB, the blue robot moves to the dock area (Fig. \ref{fig:ex1}c); as soon as the blue robot enters that room, the other robots immediately execute the instantaneous action of taking a picture (Fig. \ref{fig:ex1}d). 

\subsection{Example: Task 2}

\begin{figure}[h!]
    \centering 
    \includegraphics[width=0.75\columnwidth]{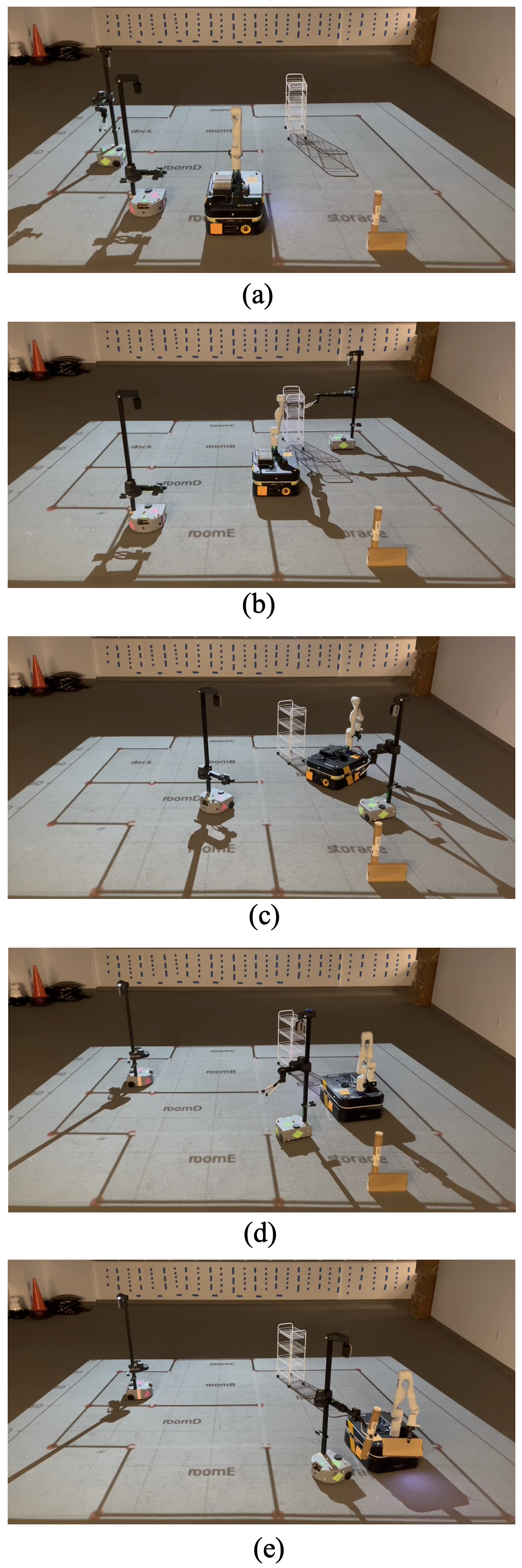}
    \caption{Robots executing task 2}
    \label{fig:ex2}
\end{figure}

For this task, we use the same environment with three available robots, $A_{green}$, $A_{orange}$, and $A_{pink}$. Their capabilities and initial positions are the same as in the previous example.

The task is 
$\varphi^\psi = \varphi^\psi_1 \wedge \varphi^\psi_2$, where
\begin{subequations} \label{ex:ex2}
    \begin{align}
    \varphi^\psi_1 &=\F \big((beep \wedge dock_c)^{1 \vee 2} \\
    &\: \wedge \F (pickup \wedge storage_c)^{1}\big) \nonumber \\ 
    \varphi^\psi_2 &= \neg roomB_c^{1 \wedge 2 \wedge 3} \ \mathcal{U} \ (push_c \wedge hall_c)^{3} 
    \end{align} 
\end{subequations}
    
    $\varphi^\psi_1$ captures ``robot(s) with bindings 1 or 2 should beep at the dock, then robot(s) with binding 1 should pickup packages at the storage. 
    $\varphi^\psi_2$ enforces ``none of the robots can be in room B until the cart in the hallway is pushed out of the way".

In this example, $push_c$, $pickup_c$, and the location completion propositions make up the set of $AP_{non-inst}$, while $beep$ is an instantaneous action. To guarantee the robots do not violate $\varphi^\psi_2$, any robot assigned binding 3 must move into the hallway first before any robots can be in room B. If we assume discrete execution of actions, this constraint is not necessary; all the robots can move into the respective areas simultaneously.   

The final team is $\hat{A}=\{$$A_{green}$, $A_{orange}$, $A_{pink}\}$ with binding assignments $r_{green} = \{1,3\}$, $r_{orange} = \{1,3\}, r_{pink} = \{2\}$.  
Snapshots of their behavior is shown in Fig. \ref{fig:ex2}. We can see that the green and orange robots first move into the hall (Fig. \ref{fig:ex2}b) to push the cart before the pink robot passes through roomB (Fig. \ref{fig:ex2}c) to make its way to the dock. The pink robot beeps immediately as it enter the dock (Fig. \ref{fig:ex2}d). Then, the green and orange robots move to pickup their respective packages at the dock and synchronize the pickup action (Fig. \ref{fig:ex2}e). 
{We assume the robots have low-level controllers to coordinate executing the $push$ and $pickup$ actions. The robots begin pushing the cart together when they have both grabbed it. The action $pickup$ includes reorienting, reaching, and grasping the packages; the robots begin executing this at the same time (i.e. they begin reorienting simultaneously).} 

\subsection{Computational Performance}

\begin{figure}[t!]
     \centering
     \begin{subfigure}{\columnwidth}
         \centering
         \includegraphics[width=0.74\textwidth]{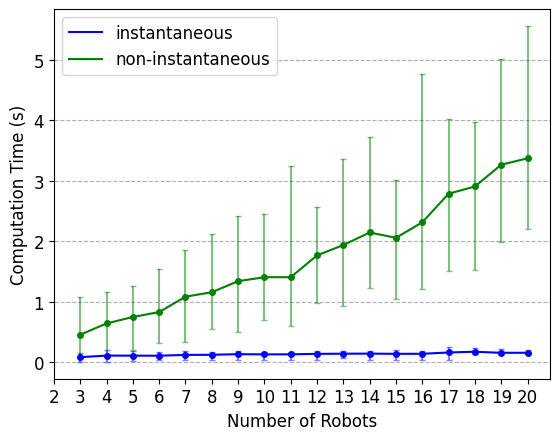}
         \caption{}
         \label{fig:vary_robots}
     \end{subfigure}
    \centering
     \begin{subfigure}{\columnwidth} 
         \centering
         \includegraphics[width=0.74\textwidth]{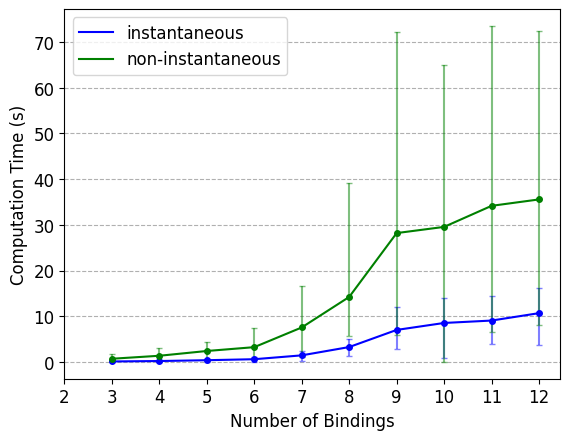}
         \caption{}
         \label{fig:vary_bindings}
     \end{subfigure}
    \caption{
    Comparing computation times between the instantaneous and non-instantaneous actions frameworks as the number of robots increases (\subref{fig:vary_robots}) and the number of bindings increases (\subref{fig:vary_bindings}). The error bars represent min/max values.}
    \label{fig:sim_results}
\end{figure}

We compare the computation performance (without significant code optimization) of the approach for synthesizing a team of robots and their corresponding binding assignments for instantaneous versus non-instantaneous actions. 

\textbf{Effects of number of robots:} We analyze the effect of increasing the number of robots while keeping the task specification the same (task 1, Eq. \ref{ex:ex1}). This task requires the generation of intermediate states when considering non-instantaneous actions. 

The instantaneous approach involves decentralized pre-processing for each robot $j$'s possible combination of binding assignments for each transition in the \buchi automaton before conducting the DFS algorithm to find an overall teaming plan. As a result, this framework is largely agnostic to the number of robots. However, the approach for non-instantaneous actions is centralized, as the DFS algorithm may need to update each robot's product automaton during each iteration. As a result, while both framework's scale linearly with the number of robots, there is a larger impact on the computational performance of the framework under the assumption of non-instantaneous actions. 

\textbf{Effects of number of bindings:} We consider the effect of increasing the number of bindings while keeping the number of robots fixed at 4, but randomizing their capabilities with each simulation. We add more bindings to the task specification through conjunction. By doing so, we increase the number of bindings, but keep the number of edges in the \buchi automaton the same.

In both the instantaneous and non-instantaneous actions frameworks, we need to store all possible binding assignments for the current team of robots as we search for a possible trace in the \buchi automaton. As a result, both the space and time complexity for each approach are exponential with the number of bindings. The non-instantaneous actions framework scales worse with the number of bindings because we determine the possible binding assignments for each robot with every transition in the DFS; in the instantaneous action framework, each robot had performed this computation in a decentralized manner before conducting the DFS algorithm.



\section{Conclusion}

We introduced a method for control synthesis for a heterogeneous multi-robot system to satisfy collaborative tasks in continuous time, where actions may take varying duration of time to complete. We presented a synthesis approach to automatically generate a teaming assignment and corresponding discrete behavior that is correct-by-construction for continuous execution and ensured collaborative portions of the task are satisfied. We implemented our approach on a physical multi-robot system in two warehouse scenarios.

In the future, we plan to explore different notions of optimality when finding a teaming plan, as well as implement methods for robots to respond to capability failures or modifications in real time. 

\bibliographystyle{ieeetr}
\bibliography{references}

\begin{thebibliography}{10}

\bibitem{Fang2024}
A.~Fang and H.~Kress-Gazit, ``High-level, collaborative task planning grammar
  and execution for heterogeneous agents,'' in {\em Proceedings of the 23rd
  International Conference on Autonomous Agents and Multiagent Systems}, AAMAS
  '24, (Richland, SC), p.~544–552, International Foundation for Autonomous
  Agents and Multiagent Systems, 2024.

\bibitem{Atay2006}
N.~Atay and B.~Bayazit, ``Mixed-integer linear programming solution to
  multi-robot task allocation problem,'' 01 2006.

\bibitem{Suslova2020}
E.~Suslova and P.~Fazli, ``Multi-robot task allocation with time window and
  ordering constraints,'' in {\em 2020 IEEE/RSJ International Conference on
  Intelligent Robots and Systems (IROS)}, pp.~6909--6916, 2020.

\bibitem{Tolmidis2013}
A.~T. Tolmidis and L.~Petrou, ``Multi-objective optimization for dynamic task
  allocation in a multi-robot system,'' {\em Engineering Applications of
  Artificial Intelligence}, vol.~26, no.~5, pp.~1458--1468, 2013.

\bibitem{Cheikhrouhou2021}
O.~Cheikhrouhou and I.~Khoufi, ``A comprehensive survey on the multiple
  traveling salesman problem: Applications, approaches and taxonomy,'' {\em
  Computer Science Review}, vol.~40, p.~100369, 2021.

\bibitem{Braekers2016}
K.~Braekers, K.~Ramaekers, and I.~{Van Nieuwenhuyse}, ``The vehicle routing
  problem: State of the art classification and review,'' {\em Computers \&
  Industrial Engineering}, vol.~99, pp.~300--313, 2016.

\bibitem{Poudel2021}
L.~Poudel, W.~Zhou, and Z.~Sha, ``{Resource-Constrained Scheduling for
  Multi-Robot Cooperative Three-Dimensional Printing},'' {\em Journal of
  Mechanical Design}, vol.~143, 04 2021.
\newblock 072002.

\bibitem{Wu2018}
W.~Wu, X.~Wang, and N.~Cui, ``Fast and coupled solution for cooperative mission
  planning of multiple heterogeneous unmanned aerial vehicles,'' {\em Aerospace
  Science and Technology}, vol.~79, pp.~131--144, 2018.

\bibitem{Holvoet2007}
T.~Holvoet, D.~Weyns, N.~Boucke, and B.~Demarsin, ``Dyncnet: A protocol for
  dynamic task assignment in multiagent systems,'' in {\em 2007 First
  International Conference on Self-Adaptive and Self-Organizing Systems}, (Los
  Alamitos, CA, USA), pp.~281--284, IEEE Computer Society, jul 2007.

\bibitem{Zhen2021}
Z.~Zhen, L.~Wen, B.~Wang, Z.~Hu, and D.~Zhang, ``Improved contract network
  protocol algorithm based cooperative target allocation of heterogeneous uav
  swarm,'' {\em Aerospace Science and Technology}, vol.~119, p.~107054, 2021.

\bibitem{Xu2015}
B.~Xu, Z.~Yang, Y.~Ge, and Z.~Peng, ``Coalition formation in multi-agent
  systems based on improved particle swarm optimization algorithm,'' {\em
  International Journal of Hybrid Information Technology}, vol.~8, pp.~1--8, 03
  2015.

\bibitem{Liu2016}
Z.~Liu, X.-G. Gao, and X.-W. Fu, ``Coalition formation for multiple
  heterogeneous uavs cooperative search and prosecute with communication
  constraints,'' in {\em 2016 Chinese Control and Decision Conference (CCDC)},
  pp.~1727--1734, 2016.

\bibitem{Luo2022}
X.~Luo and M.~M. Zavlanos, ``Temporal logic task allocation in heterogeneous
  multirobot systems,'' {\em IEEE Transactions on Robotics}, pp.~1--20, 2022.

\bibitem{Sahin2017}
Y.~E. Sahin, P.~Nilsson, and N.~Ozay, ``Synchronous and asynchronous
  multi-agent coordination with cltl+ constraints,'' in {\em 2017 IEEE 56th
  Annual Conference on Decision and Control (CDC)}, pp.~335--342, 2017.

\bibitem{Chen2021}
J.~Chen, R.~Sun, and H.~Kress-Gazit, ``Distributed control of robotic swarms
  from reactive high-level specifications,'' in {\em 2021 IEEE 17th
  International Conference on Automation Science and Engineering (CASE)},
  pp.~1247--1254, 2021.

\bibitem{Leahy2022}
K.~Leahy, Z.~Serlin, C.-I. Vasile, A.~Schoer, A.~M. Jones, R.~Tron, and
  C.~Belta, ``Scalable and robust algorithms for task-based coordination from
  high-level specifications (scratches),'' {\em IEEE Transactions on Robotics},
  vol.~38, no.~4, pp.~2516--2535, 2022.

\bibitem{Gundana2021}
D.~Gundana and H.~Kress-Gazit, ``Event-based signal temporal logic synthesis
  for single and multi-robot tasks,'' {\em IEEE Robotics and Automation
  Letters}, vol.~6, no.~2, pp.~3687--3694, 2021.

\bibitem{Buyukkocak2021}
A.~T. Buyukkocak, D.~Aksaray, and Y.~Yazıcıoğlu, ``Planning of heterogeneous
  multi-agent systems under signal temporal logic specifications with integral
  predicates,'' {\em IEEE Robotics and Automation Letters}, vol.~6, no.~2,
  pp.~1375--1382, 2021.

\bibitem{Tumova2016}
J.~Tumova and D.~V. Dimarogonas, ``Multi-agent planning under local ltl
  specifications and event-based synchronization,'' {\em Automatica}, vol.~70,
  pp.~239--248, 2016.

\bibitem{Liu2023}
W.~Liu, K.~Leahy, Z.~Serlin, and C.~Belta, ``Robust multi-agent coordination
  from catl+ specifications,'' 2023.

\bibitem{Raman2012}
V.~Raman, C.~Finucane, and H.~Kress-Gazit, ``Temporal logic robot mission
  planning for slow and fast actions,'' in {\em 2012 IEEE/RSJ International
  Conference on Intelligent Robots and Systems}, pp.~251--256, 2012.

\bibitem{Moarref2017}
S.~Moarref and H.~Kress-Gazit, ``Decentralized control of robotic swarms from
  high-level temporal logic specifications,'' in {\em 2017 International
  Symposium on Multi-Robot and Multi-Agent Systems (MRS)}, pp.~17--23, 2017.

\bibitem{EMERSON1990}
E.~A. Emerson, ``Temporal and modal logic,'' in {\em Formal Models and
  Semantics} (J.~{Van Leeuwen}, ed.), Handbook of Theoretical Computer Science,
  pp.~995--1072, Amsterdam: Elsevier, 1990.

\bibitem{Baier2008}
C.~Baier and J.-P. Katoen, {\em Principles of Model Checking}.
\newblock The MIT Press, 2008.

\bibitem{Fang2022}
A.~Fang and H.~Kress-Gazit, ``Automated task updates of temporal logic
  specifications for heterogeneous robots,'' in {\em 2022 International
  Conference on Robotics and Automation (ICRA)}, pp.~4363--4369, 2022.

\bibitem{spot}
A.~Duret-Lutz, A.~Lewkowicz, A.~Fauchille, T.~Michaud, E.~Renault, and L.~Xu,
  ``Spot 2.0 --- a framework for {LTL} and $\omega$-automata manipulation,'' in
  {\em Proceedings of the 14th International Symposium on Automated Technology
  for Verification and Analysis (ATVA'16)}, vol.~9938 of {\em Lecture Notes in
  Computer Science}, pp.~122--129, Springer, Oct. 2016.

\end{thebibliography}
\end{document}